\documentclass[outdir=./]{article}
\usepackage{frExamplee}
\usepackage[natbibapa]{apacite}
\usepackage{setspace}
\usepackage{amsmath}
\usepackage{multirow}
\usepackage{lscape}
\usepackage{graphicx}
\usepackage{algorithm}
\usepackage{algpseudocode}
\usepackage{epstopdf}
\usepackage{epstopdf}
\usepackage{mathtools}
\usepackage{amsmath,amssymb}

\usepackage{caption} 
\usepackage{nameref}
\usepackage{pifont}

\usepackage{adjustbox}
\usepackage{hyperref}
\hypersetup{
    colorlinks=true,
    linkcolor=blue,
    filecolor=magenta,      
    urlcolor=cyan,
    pdftitle={Overleaf Example},
    pdfpagemode=FullScreen,
}
\usepackage{svg}
\usepackage{subcaption}
\usepackage{cleveref}
\usepackage{wrapfig}

\begin{document}

\title{An Autonomous System for Head-to-Head Race: 
Design, Implementation and Analysis;\\ Team KAIST at the Indy Autonomous Challenge}

\author{
Chanyoung Jung\thanks{These authors contributed equally to this work.} , \\
KAIST \\
\texttt{cy.jung@kaist.ac.kr} \\
\And
Andrea Finazzi$^*$,\\
KAIST \\
\texttt{finazzi@kaist.ac.kr} \\
\And
Hyunki Seong,\\
KAIST \\
\texttt{hynkis@kaist.ac.kr} \\
\And
Daegyu Lee,\\
KAIST \\
\texttt{lee.dk@kaist.ac.kr} \\
\And
Seungwook Lee,\\
KAIST \\
\texttt{seungwook1024@kaist.ac.kr} \\
\And
Bosung Kim,\\
KAIST \\
\texttt{brian.kim@kaist.ac.kr} \\
\And
Gyuri Gang,\\
KAIST \\
\texttt{fingb20@kaist.ac.kr} \\
\And
Seungil Han,\\
KAIST \\
\texttt{robotics@kaist.ac.kr} \\
\And
David Hyunchul Shim\thanks{David Hyunchul Shim is the corresponding author.} \\
KAIST \\
\texttt{hcshim@kaist.ac.kr} \\
}

\maketitle

\begin{abstract}
While the majority of autonomous driving research has concentrated on everyday driving scenarios, further safety and performance improvements of autonomous vehicles require a focus on extreme driving conditions. In this context, autonomous racing is a new area of research that has been attracting considerable interest recently. Due to the fact that a vehicle is driven by its perception, planning, and control limits during racing, numerous research and development issues arise. This paper provides a comprehensive overview of the autonomous racing system built by team \emph{KAIST} for the Indy Autonomous Challenge (IAC). Our autonomy stack consists primarily of a multi-modal perception module, a high-speed overtaking planner, a resilient control stack, and a system status manager. We present the details of all components of our autonomy solution, including algorithms, implementation, and unit test results. In addition, this paper outlines the design principles and the results of a systematical analysis. Even though our design principles are derived from the unique application domain of autonomous racing, they can also be applied to a variety of safety-critical, high-cost-of-failure robotics applications. The proposed system was integrated into a full-scale autonomous race car (Dallara AV-21) and field-tested extensively. As a result, team \emph{KAIST} was one of three teams who qualified and participated in the official IAC race events without any accidents. Our proposed autonomous system successfully completed all missions, including overtaking at speeds of around 220 ${km}/{h}$ in the IAC@CES2022, the world's first autonomous 1:1 head-to-head race.
\end{abstract}

\section{Introduction}
\label{sec:introduction}
The Society of Automotive Engineers (SAE) defines level 5 autonomy as the full-time performance of the driving task without human intervention by an autonomous driving system. The potential advantages of autonomous vehicles are immense. According to a report by the consulting firm McKinsey and Company, autonomous vehicles will help reduce annual traffic fatalities in the United States by up to 90\%, save commute time for high productivity, decrease traffic congestion and pollution, and boost the utilization of driving resources \citep{WinNT}.
In parallel with the development of hardware technology, autonomous driving software has been a focus of intense interest among academia and industry during the past two decades. As a result, modern Advanced Driver Assistance Systems (ADAS) systems of Level 3 or highly automated prototypes are readily accessible, and legislation mandating the adoption of ADAS on everyday vehicles is being pushed in certain countries \citep{brodsky2016autonomous}.

\begin{figure}[b!]
\centerline{\includegraphics[width=1.0\textwidth]{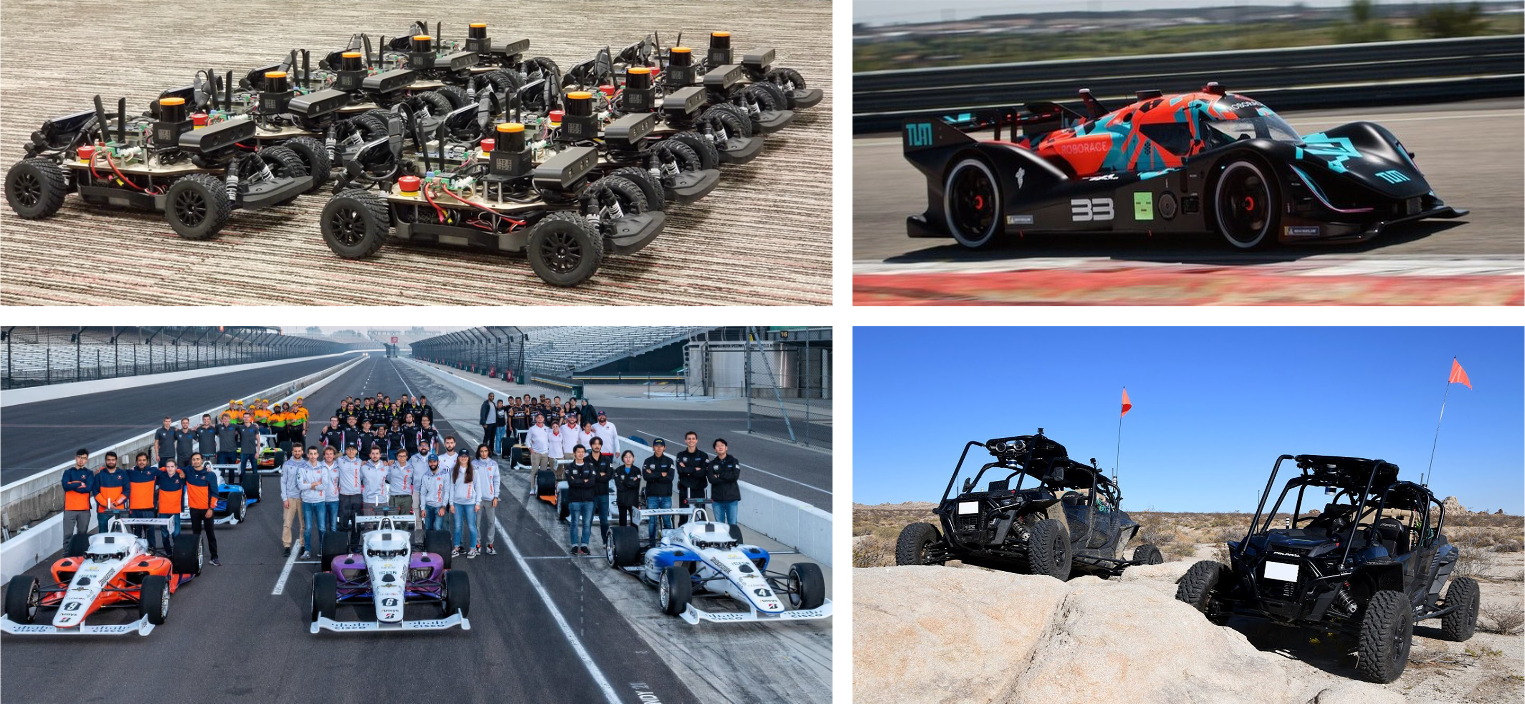}}
\caption[]{Examples of real-world autonomous racing competitions. (Top-Left) F1 TENTH, using a 1:10 scaled vehicle. (Top-Right) ROBORACE, using a full-scale electric race car. (Bottom-Left) Indy Autonomous Challenge (IAC), using a Dallara AV-21 retrofitted Indy Lights class chassis. (Top-Left) DARPA-RACER program, using autonomous ground combat vehicles in unstructured off-road terrain at speeds.} \label{fig:background_other_racing_compeitions}
\end{figure}

To achieve further safety, reliability, and performance improvements of autonomous driving technology, the autonomous racing field of research is gaining a lot of attention these days \citep{betz2022autonomous}. This is the same innovation pathway that links Formula 1 to our everyday vehicles. Driven by this growing interest, several real-world autonomous race events such as F1TENTH \citep{o2020f1tenth}, Roborace \citep{rieber2004roborace}, Indy Autonomous Challenge \citep{iac}, and Darpa-RACER \citep{racer} have been held (See Fig. \ref{fig:background_other_racing_compeitions}).

Autonomous racing poses numerous research and development challenges since a vehicle is pushed to its perception, planning, and control limits \citep{talvala2011pushing,betz2022autonomous}. For example, an autonomous race car should be able to detect opponents from a far distance and accurately predict future trajectories based on multiple factors, including environment, strategy, and race rules, in order to safely overtake. Furthermore, as professional human drivers do, autonomous racing vehicles should be able to plan their trajectory and control the vehicle at high speeds considering vehicle dynamics. Most challenging, however, is that the autonomous system should fulfill the combination of the above requirements with limited computational resources and compute everything in real time. 

In this paper, we present the team \emph{KAIST}'s autonomy solution for the Indy Autonomous Challenge, the world' first head-to-head racing competition featuring full-scale, autonomous racing vehicles (described in Section \ref{sec:iac}). Our paper is organized into three major thematic sections: 1. Principles of the overall system design 2. A full-stack software for autonomous racing 3. Results of extensive field testing and in-depth system performance analysis. Outlining our design principles is not only helpful to understand the proposed autonomy architecture, but also to give insight into a wide range of field robotics applications. Following the design principles, we built a full-stack autonomous racing software composed of a multi-modal perception module, a high-speed overtaking planner, and a resilient control stack. Every subsystem was evaluated in the context of the autonomous racing competition. Our software stack was fully integrated into the Dallara-AV21 which served as the IAC's official vehicle platform and extensively tested in the real-world. Furthermore, we deliver the result of an in-depth computational performance analysis. 

Autonomous racing has a lot of technical challenges, which come from the fact that various algorithms, including perception, planning, and control are pushed to the limits. Furthermore, system design also possesses considerable challenges. A complex autonomy stack should operate in real time within limited computing resources. Our study addressed these challenges and intensively evaluated the entire system based on real-world experiments using a full-scaled race vehicle. Contributions of this paper are not limited to the field of autonomous driving racing. We believe that our autonomy design and real-world evaluation results provide technical insight into research on various autonomous robot applications operating in extreme environments.

In summary, the contributions of this paper are the followings:
\begin{enumerate}
\item We propose a full-stack autonomous racing system that includes multi-modal perception, high-speed overtaking planner, and resilient control stack.
\item We integrate the proposed autonomy solution into  a full-scaled autonomous race vehicle (Dallara AV-21). The result shows that our system has capabilities by reaching speeds of over 220 ${km}/{h}$ and lateral and longitudinal accelerations of up to 6.8 ${m}/{s^2}$ and 12.4 ${m}/{s^2}$, respectively.
\item We evaluate the performance of every modules of our autonomy stack in the context of high-speed/head-to-head autonomous racing.
\item We present the system's design principles and an in-depth computational performance analysis. Even though our design principles are derived from the field of autonomous racing, we believe that these can be applied to various field robotics/safety-critical applications.
\end{enumerate}

The paper is organized as follows: Section \ref{sec:relatedwork} presents the previous literature on the field of agile/autonomous racing. 
Section \ref{sec:iac} overview of the IAC including rules, timelines and the spec of official racing fleet. 
The design principles are described in section \ref{sec:design-principles}. The multi-modal perception pipeline, the high-speed overtaking planner, and the resilient control stack are covered in section \ref{sec:autonomous-racing-system} with a brief introduction of the official race vehicle platform, Dallara-AV21. The field testing results and the computational performance analysis are provided in section \ref{sec:evaluation}. Finally, section \ref{sec:conclusion} summarizes the conclusions from this two year project and future plans for the next seasons.

\section{Related Work}
\label{sec:relatedwork}
There are numerous existing studies relating to autonomous driving systems comprising perception, decision/planning, and control modules. This section gives an overview of previous works on the core tasks of autonomous racing systems.

\subsection{Agile Control}
\label{sec:agile-control}
Model Predictive Control (MPC) is a widely and well-studied controller for high-speed autonomous vehicles \citep{carvalho2013predictive,funke2015collision,rosolia2019learning}. Even though we classified MPC under the control category, a benefit of the optimization approach is that it combines some of the traditionally separate planning and control modules by creating one optimization framework that operates on a task description and dynamics model of the system. In the work of \citep{liniger2015optimization}, the model predictive contouring control (MPCC), which can track a given reference path, is adapted for autonomous driving applications. The authors added a contouring cost term which represents the tracking error to the objective function. They demonstrated high-speed driving and collision avoidance experiments using a 1:43 scaled vehicle. \citep{goldfain2019autorally} conducted rally racing in an outdoor, dirt environment using the Model Predictive Path Integral control (MPPI) framework, a stochastic optimal controller. The authors modeled the nonlinear vehicle dynamics using a simple neural network and trained the model in a supervised manner. They randomly rolled out the vehicle's future states repetitively and generated the optimal control signal via information theory. They integrated the proposed framework into a 1:5 scaled offroad vehicle platform and tested it in an unpaved road environment. However, the performance of model-based approaches is directly affected by the accuracy of the models. Finding the model parameters is not trivial, and the result is necessarily an approximation. Thus, most previous studies simplified the optimization problem using linearized models.

With the development of machine learning technologies, recently, a number of studies have demonstrated the success of using model-free deep RL for autonomous driving, and racing \citep{jaritz2018end,riedmiller2007learning,kendall2019learning,grigorescu2020survey,cai2020high,wurman2022outracing}. \citep{fuchs2021super} demonstrated high-speed autonomous driving using model-free RL within a high-fidelity simulation environment by utilizing a course-progress proxy reward. They argue that because their model directly outputs the control command (one-step RL), it has the advantage of not relying on high-level trajectory planning and following while generating trajectories qualitatively similar to those chosen by the best human drivers.
Even though RL approaches show promising results in the field of control, it still poses a lot of technical and practical challenges when it comes to the real-world field robotics domain. For example, RL based policy can not be clearly validated before deployment. Especially, one-step RL setup makes hard to incorporate with the other navigation and control algorithms which leads the lack of system resilience. 

\subsection{Planning for Racing}
\label{sec:competitive-planning}
Given the vehicle and track models, minimum lap time or time-optimal trajectory planning has been studied in various automation fields. Multiple algorithms have been studied to create an optimal line focusing on lap-time or fuel consumption depending on their target applications. An attempt was made to develop it based on data from professional race car drivers in the racing field with the sole purpose of achieving minimum lap time \citep{milliken1995race}. The analysis result of the driving trajectory was used to shorten the lap time and design the vehicle. In early 2000, \citep{casanova2000minimum} proposed a Nonlinear Programming (NLP)-based method which optimizes the path and velocity profile simultaneously using a nonlinear vehicle dynamics model.
Another widely used approach is to convert the minimal lap time problem into an MPC problem \citep{timings2013minimum,liniger2015optimization}. They all created a nonlinear model-predictive framework to solve an optimal control problem with time as the objective. As an alternative, the shortest lap time problem can be approximated by minimizing the lateral acceleration problem, leading to finding the race line with the minimum curvature. Of course, because it does not explicitly tackle the optimization problem of lap time, this solution cannot ensure global optimality. It can, however, be useful when the precise vehicle dynamics and nonlinear tire model parameters are unknown. \citep{heilmeier2019minimum} used quadratic programming to compute a minimum-curvature racing line, and simulation testing revealed that it performs fairly similarly to the minimum-lap time. This method has the advantage of not requiring sophisticated vehicle model parameters.

Local trajectory planning for high-speed vehicles is a module that plans a non-colliding and dynamically feasible fixed horizon of trajectory near a given global race line. MPC is mainly used in the optimal control field, but it is also used as a path planner through state propagation using a model and optimal control output \citep{williams2016aggressive,subosits2019racetrack,funke2016collision}. Another category for local path planning is using motion primitives in a more general control-space sampling-based planner. This approach generates multiple candidates by propagating the model given the vehicle's current state and selecting the best one based on the designed cost function. \citep{liniger2015viability} generate a library of trajectories by forward-simulating the vehicle using a grid of vehicle velocities and steering angles up to a certain horizon. The final local trajectory was chosen by minimizing the curvature while traversing the track to maximize the velocity and keep the vehicle as straight as possible. By adjusting the size of the sampling space, this approach presents a trade-off between optimality and computational burden. To leverage the fact that the track geometry is fixed, \citep{stahl2019multilayer} proposed a two steps sampling-based local path planner for a race vehicle. In an offline process, they build the track as a graph consisting of layers, nodes, and edges. The connectivity of node and edge is determined based on the kinematic constraints of the vehicle. Then, a cost is assigned to each edge according to the displacement from the race line and the curvature. The local path is planned in the online process by performing a minimum cost path search of a fixed time horizon branch based on the track graph model. This planner was integrated into a Roborace vehicle and validated over the 200 ${km}/{h}$ speed range.

\subsection{Autonomous Race Vehicles and Competitions}
\label{sec:autonomous-race-vehicle-competition}
The majority of research on autonomous racing has been conducted utilizing simulation, or scaled vehicle platforms \citep{herman2021learn,weiss2020deepracing,liniger2017real,kabzan2020amz}. Exploiting advanced physics engine and graphic technology, simulation enables a variety of experiments that would be difficult to carry out in the real world. Also, considering the cost of operation and safety, the scaled platform has been employed in several researches. Unfortunately, there is still a gap between simulation and reality, making simulation impossible to estimate the algorithm's scalability precisely. This section will concentrate on a full-scaled autonomous vehicle and competitions.
\begin{figure}[t!]
\centerline{\includegraphics[width=0.97\textwidth]{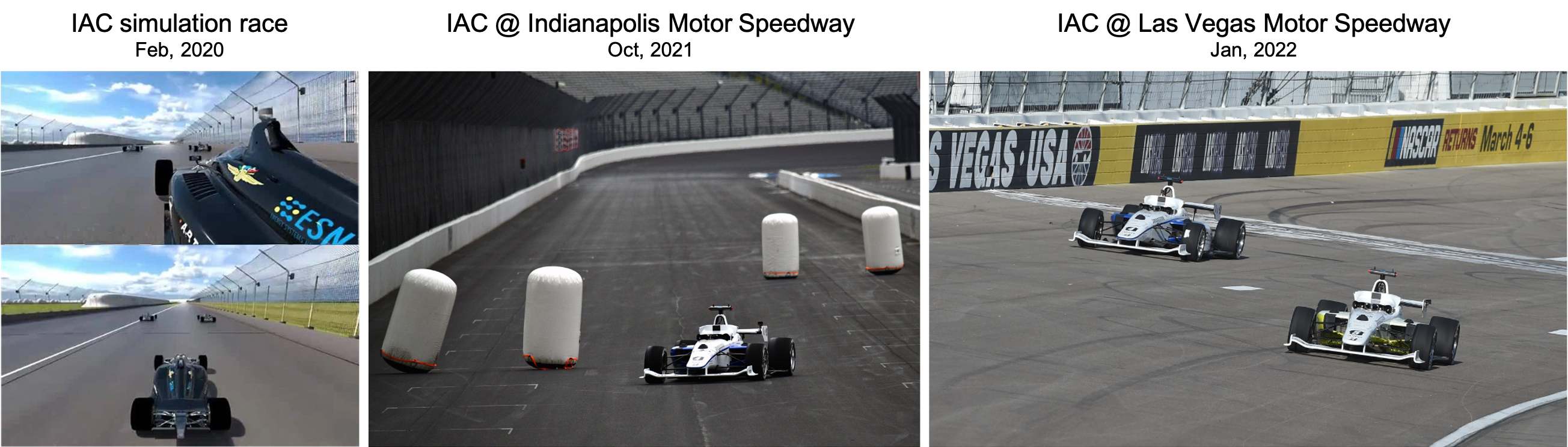}}
\caption[Overview of the timeline of IAC.]{Overview of the IAC timeline. (Left) Simulation phase (Middle) The first real-world high-speed competition at IMS (Right) 1:1 head-to-head autonomous race event at LVMS.} \label{fig:iac_all}
\end{figure}

In collaboration with Stanford University, Audi debuted Shelly, an autonomous TTS capable of high-speed autonomous driving \citep{funke2012up}. They reached a top speed of about 305 ${km}/{h}$ at the Salt Flats in Utah, and high-speed autonomous driving at the Pikes Peak International Hill Climb in 2009. Since they focused on time trial race or high-speed driving, it is only equipped with an integrated Differential Global Positioning System (DGPS) and Inertial Measurement Unit (IMU) as sensors. 

The Roborace platform is based on an electric vehicle designed for the autonomous racing competition \citep{rieber2004roborace}. The platform is based on a Le Mans Prototype chassis and is equipped with cameras, LiDARs, and radars. Starting in 2018, the competition was held as a season event, with time trials. There were virtual obstacles on the track, and each team should develop a software stack aiming to complete the race mission without any time penalty. Vehicles are not provided for each team but rather in a way that allows their software to be deployed across vehicles.

The Indy Autonomous Challenge (IAC) is the most recent autonomous racing competition. The Dallara-AV21, the official vehicle platform for the competition, was constructed on an Indy Lights chassis and has a combustion engine. More information about the race will be provided in the section that follows.

\section{Indy Autonomous Challenge}
\label{sec:iac}
As shown in Fig. \ref{fig:iac_all}, IAC began in November 2019. It was divided into several stages: document screening, hackathons, simulation races, and real-world races. About 30 teams participated in the simulation race, and 9 of them, including our team, competed in the real-world autonomous race competition. The first real-world competition held at Indianapolis Motor Speedway (IMS) in 2021 and was conducted with a time trial racing format. Teams had to complete pit-in, pit-out, performance lap, and static avoidance tasks following the race control signals. The next race, part of the official CES 2022 event, was held in a 1:1 head-to-head race format, with two vehicles racing on the course simultaneously. Several artificial rules were imposed: 1. The defending vehicle must only run on the inner line of the track at the commanded speed. 2. The defending and attacking vehicles must adhere to race control, and overtaking is restricted to a specific region and race flag. 3. The attacking vehicle should close the overtaking maneuver when a safety distance of 20m is secured. 

\begin{figure}[b!]
\centerline{\includegraphics[width=.85\textwidth]{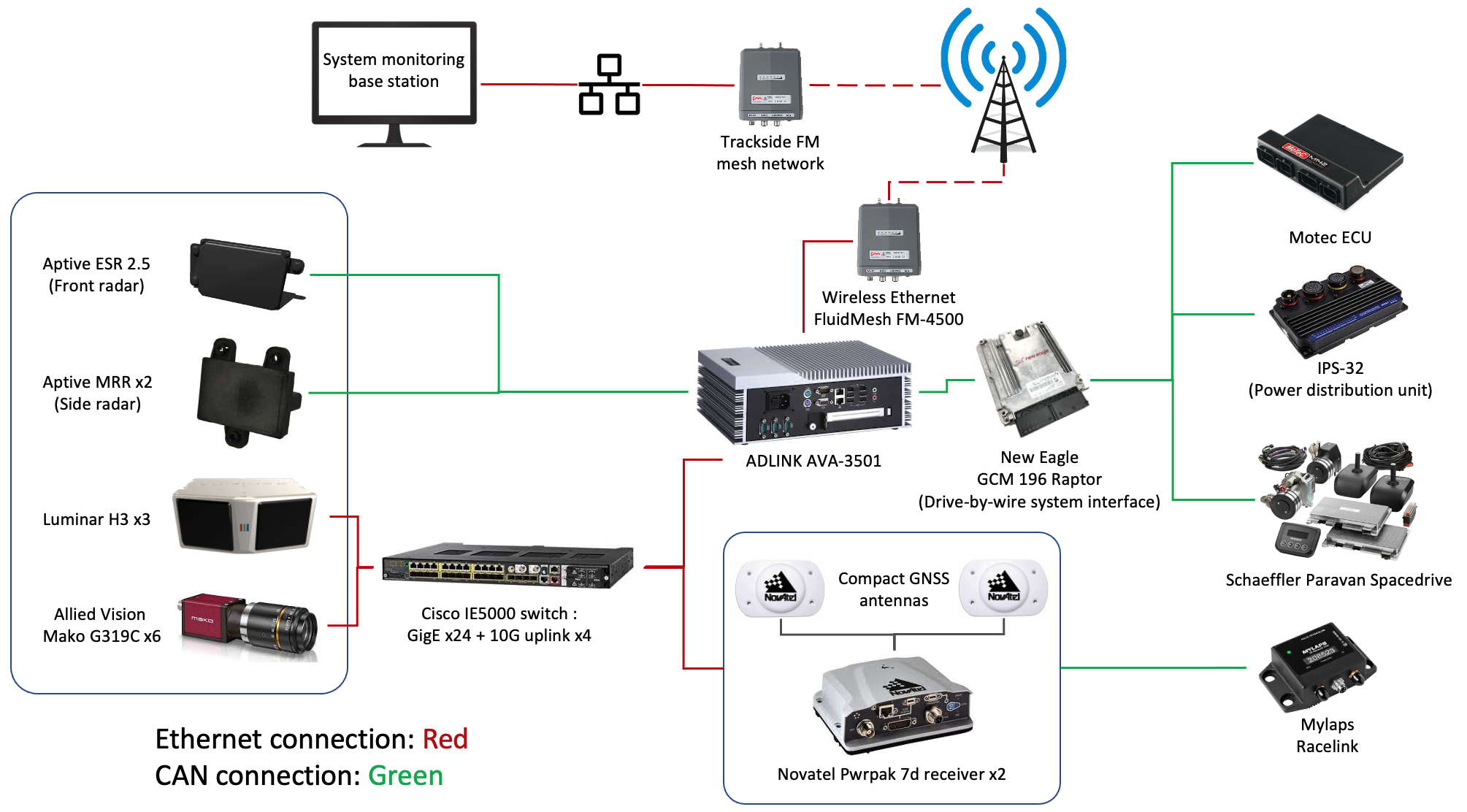}}
\caption[System diagram of the Dallara AV-21.]{System diagram of the Dallara AV-21.} \label{fig:system_diagram_}
\end{figure}

The participating teams used the same vehicle platform, the Dallara AV-21, adapted for autonomous racing, and they had to design and integrate their own autonomy stack. The vehicle is rear-wheel drive, powered by an internal combustion engine which produces 335kW (449hp) and has a 6-speed sequential gearbox. Computing devices, sensors, and controllers were placed instead of the driver seat. Six Gig-E cameras, three Radars, three solid-state LiDARs, and an RTK GPS are equipped as a sensor package. The computing platform included an Intel Xeon CPU with an Nvidia Quadro RTX 8000 GPU. Fig. \ref{fig:system_diagram_} shows the system diagram of the Dallara AV-21.

\section{Design Principles}
\label{sec:design-principles}
Autonomous racing poses a few challenges in addition to the functional requirements of an urban-oriented autonomous driving system. The capabilities of the platform in terms of acceleration and maximum speed, along with the particular environment that are racing tracks, make the application domain unique. Moreover, the pioneering nature of the IAC competition introduces a high level of uncertainty in the definition of the functional requirements. 

\begin{figure}[t!]
\centerline{\includegraphics[width=0.8\textwidth]{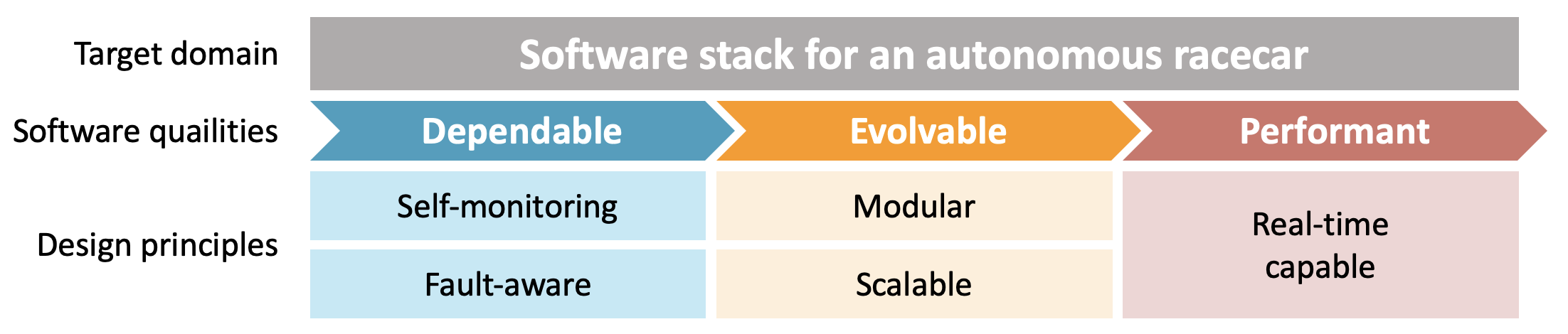}}
\caption[Key design principles of team \emph{KAIST} autonomy stack.]{Key design principles of team \emph{KAIST} autonomy stack.} \label{fig:design_principles}
\end{figure}

To cope with the above-mentioned challenges, dependability, evolvability, and performance were identified as the highest-priority software qualities at the early stages of the design process. Our key design principles are shown in Fig. \ref{fig:design_principles}.
Dependability is the measure of the trustworthiness of a software system. \citep{sommerville2015se10} describes it as a five-dimensional quality. Being autonomous racing a safety-critical and high cost-of-failure application, we define dependability as an aggregate measure of reliability, safety, and resilience, as availability and security are of secondary importance in this work's application domain. 
Evolvability refers to the ability of a system to respond to functional requirements modifications, handle domain uncertainty, and absorb change without incurring disruption. This software quality is an important aspect of field-robotics systems since the requirements of the problem can change according to various factors, such as environmental conditions, field testing scenarios, and so on. 
Lastly, performance; this can be defined in several ways, and different classifications of performance and real-time systems are available in the literature \citep{Oshana2006OverviewSystems}. In our work, performance is intended as the capability of the system to respect execution deadlines and operation frequencies defined at the design stage, analogously to what is asked to a soft real-time system.

Out of these properties, the following requirements were defined and adopted throughout the development of the system:
\begin{itemize}
    \item \textbf{Self-monitoring}: the system has the ability to detect and react to potential failures at different levels (infrastructure, application, platform).
    \item \textbf{Fault-aware}: the potential occurrence of faults (either internal or external) is taken into account as a fundamental modeling principle. Clear separation of duties and a fail-fast \citep{shore2004failfast} development approach were adopted to satisfy this requirement.
    \item \textbf{Modular}: adding, removing, and exchanging functional components with minimal integration effort is necessary to respond effectively to software evolution and functional requirements uncertainty.
    \item \textbf{Scalable}: the architecture as a whole, as well as the single components, must be able to grow as the system's complexity increases and new functionalities are introduced.
    \item \textbf{Real-time capable}: execution deadlines and operation frequency have to be defined and respected by the software components, as in an embedded soft real-time system.
\end{itemize}

\section{Autonomous Racing System}
\label{sec:autonomous-racing-system}

\subsection{Architecture Overview}
\label{sec:architecture-overview}
\begin{figure}[t!]
\centerline{\includegraphics[width=1.\textwidth]{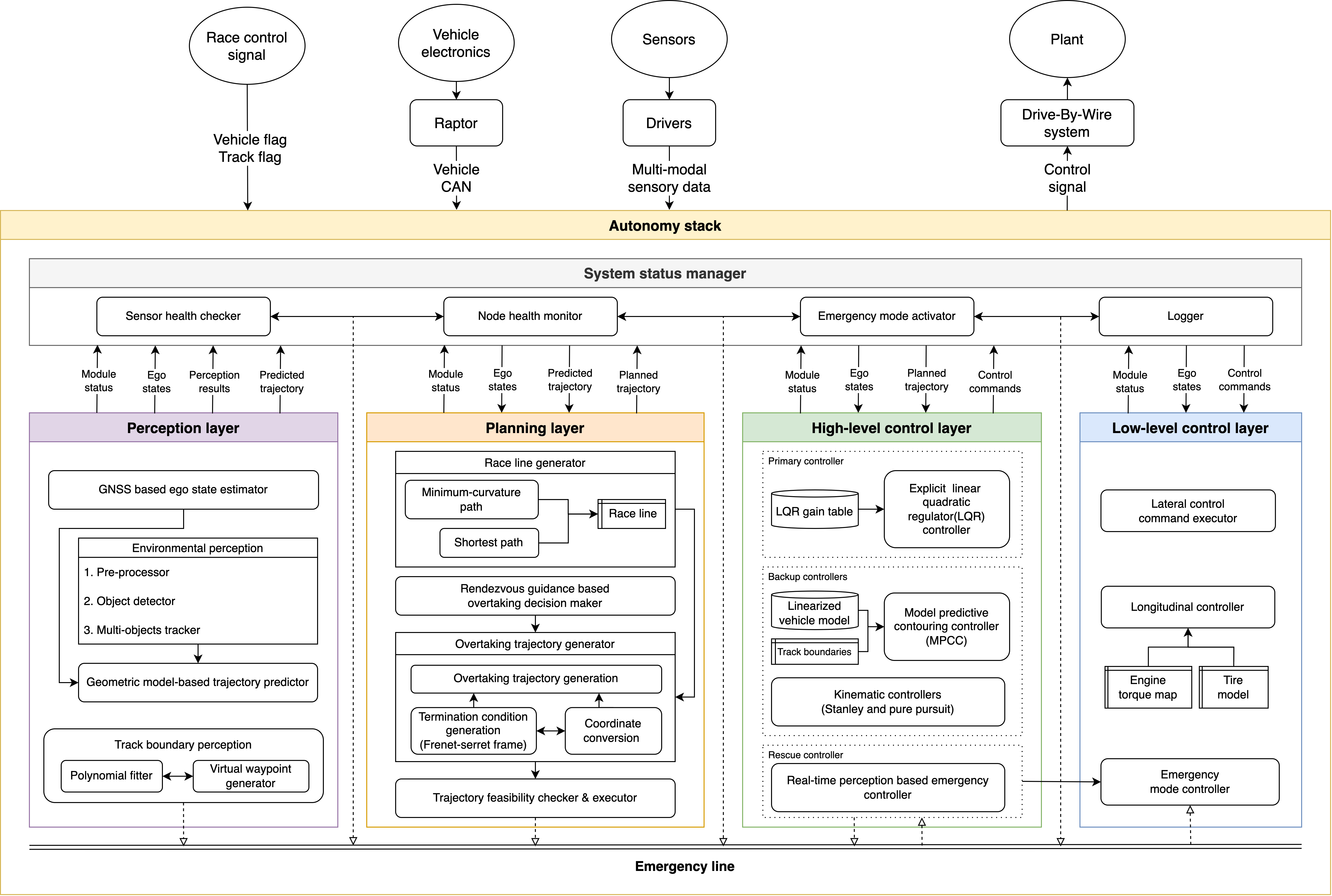}}
\caption[Overview of team \emph{KAIST}'s software architecture.]{Overview of team \emph{KAIST}'s software architecture.} \label{fig:overall}
\end{figure}

Fig. \ref{fig:overall} depicts a high-level functional block diagram of the proposed autonomous racing system. Our autonomy stack is designed to address the head-to-head race scenarios. The system comprises several subsystems: 1. System status manager; 2. Perception layer; 3. Planning layer, and 4. High-level and low-level control layers. Every subsystem is developed following our design principles with different requirements (e.g., update frequency, type of output).

As previously stated, our autonomy stack follows the traditional perception-planning-control task flow. We designed a multi-modal perception pipeline based on the equipped sensors of the Dallara-AV21. The perception module's outputs are fed into the planning module. Then, the planning module is responsible for generating a collision-free trajectory that can also overtake the other opponent. Finally, the control subsystem computes the desired lateral and longitudinal control commands to precisely follow the planned trajectory while considering vehicle dynamics.
Furthermore, our control stack is designed to support multiple control algorithms running in parallel to cope with system failures. Besides the functional subsystems (perception, planning, and control layers), a system status manager was designed to ensure system resilience by monitoring the health of each subsystem. For instance, if any module is not in the nominal status, the system status manager dynamically reacts at a system level to recover or safely stop the vehicle. More information regarding our system will be provided in the following sections.

\subsection{System Status Manager}
\begin{figure}[t!]
    \centerline{\includegraphics[width=0.84\textwidth]{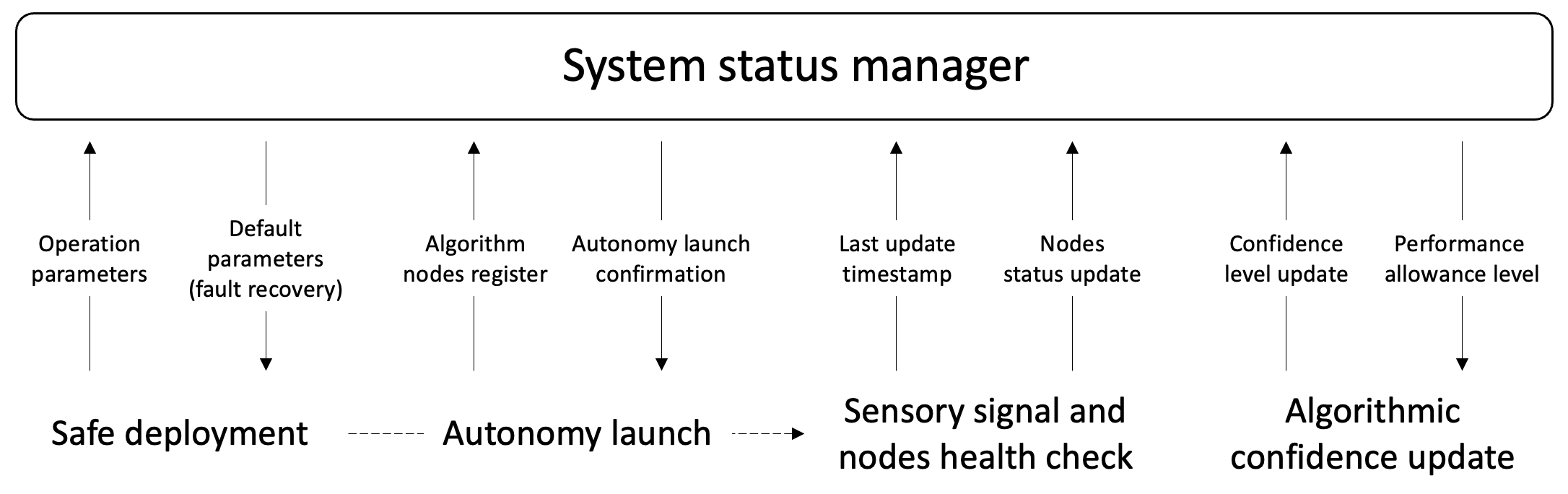}}
    \caption[Overview of the behavior of system status monitor by operation stage.]{Overview of the behavior of system status monitor by operation stage.} 
    \label{fig:ssm}
\end{figure}

Autonomous racing is a safety-critical and high-cost of failure application. As a result, a fully automated mechanism is necessary to detect system abnormalities and initiate appropriate recovery behavior. With that aim, we have configured a system status manager (hereafter referred to as SSM) that supervises the state of submodules that compose the autonomy stack. As shown in Fig. \ref{fig:ssm}, our SSM contributes to the system's resilience through different criteria according to the operation phase.

The SSM starts by checking parameters in the operation configuration to ensure safe deployment. Multiple parameters must be pre-set according to the testing purpose and race operation strategy. However, manually inputted parameters by the user may be incorrect and lead to significant performance degradation or fatal accidents during vehicle operation. To prevent faulty operation by human operators, our SSM validates the parameters during the deployment/launching phase, such as pre-set value range, sign, and data type. Representative parameters managed by the SSM include maximum speed, maximum acceleration/deceleration, time-out threshold for sensors, and watchdog threshold. In case any parameter does not meet the conditions, the SSM immediately suspends the autonomy launch and applies full braking. 

\begin{wraptable}{r}{0.45\textwidth}
\caption[Pre-set node status list and codes.]{Pre-set node status list and codes.}
\label{tab:node_status_list}
\centering
\begin{tabular}{cc}
\hline
Node status name  & Code \\ \hline \hline
NODE\_OK               & 0             \\ \hline
NODE\_INITIALIZED      & 64            \\ \hline
NODE\_INACTIVE         & 100           \\ \hline
NODE\_NOT\_INITIALIZED & 128           \\ \hline
NODE\_ERROR            & 200           \\ \hline
NODE\_DEAD             & 255           \\ \hline
\end{tabular}
\end{wraptable}
After the SSM confirms that configurations are OK, every algorithm module is launched. During the online autonomy phase, the SSM periodically checks the health of sensory signals. Here, we consider the update rate and the size of the sensor data as health indicators. The SSM manages all the status of sensory data centrally, and every algorithm module which utilizes the raw sensor data was strictly designed to check its health before the algorithmic computation. Similarly, all algorithm blocks (hereafter referred to as nodes) that compose the autonomy stack report the node status to the SSM. As shown in Table \ref{tab:node_status_list}, the state of each node is updated based on its phase. As a result, the SSM determines that the current system is under a nominal situation only when all registered node status codes are OK (code 0). The SSM responds differently depending on the type of error and on the node where the error occurs.

To monitor the system's state more precisely, the nodes communicate the reliability of algorithm's result during operation. Measuring the reliability or uncertainty of solutions is one of the key aspects of designing a resilient system. Our idea was to cope with abnormality by using the degraded mode of operation. For example, when our system is not confident about the GNSS-based localization result, our SSM lowers the maximum speed, acceleration, and so on. Furthermore, when the localization node incurs into fatal error at high-speed driving, the SSM switches to a controller that does not rely on GNSS-based location signals. 
Details about our efforts for monitoring the system status will be provided in the according sections. 

\subsection{Multi-modal Perception Pipeline}
\label{sec:perception}
Estimating the ego's state and understanding the surroundings are the first steps of the autonomous driving task. Perception performance directly impacts the system's capability and its overall robustness. Unique features of the perception system for autonomous racing are as follows: 1. Since the race vehicle drives only in a known and controlled environment (i.e., a race track), the perception system can fully utilize its geometric information. 2. It can be reasonably assumed that only race vehicles exist on the track in most cases. 
3. Considering the driving speed, the perception range and the detection update rate should be long and fast enough to safely operate. 

\begin{figure}[t!]
\centerline{\includegraphics[width=0.98\textwidth]{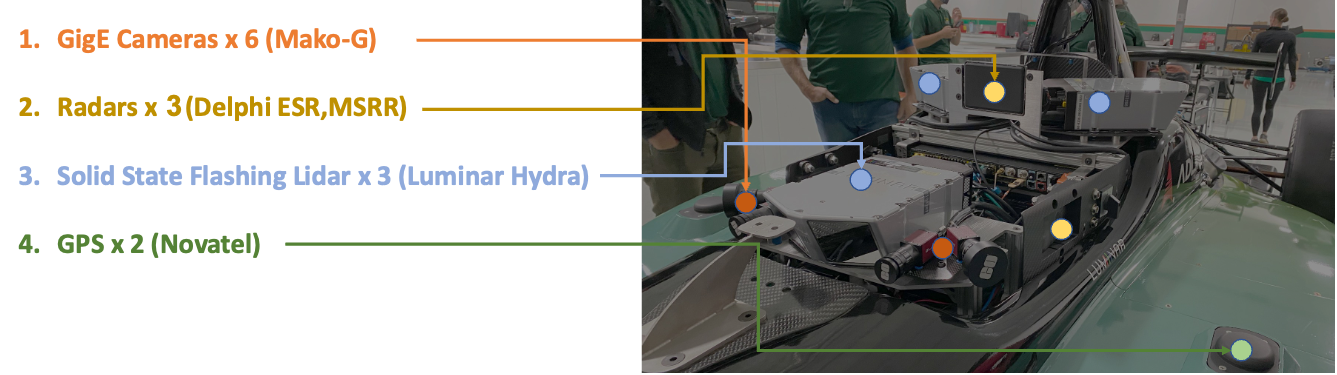}}
\caption[Dallara AV-21 sensor configuration.]{Dallara AV-21 sensor configuration.} \label{fig:sensors_layout_realveh}
\end{figure}

\begin{figure}[t!]
\centerline{\includegraphics[width=0.98\textwidth]{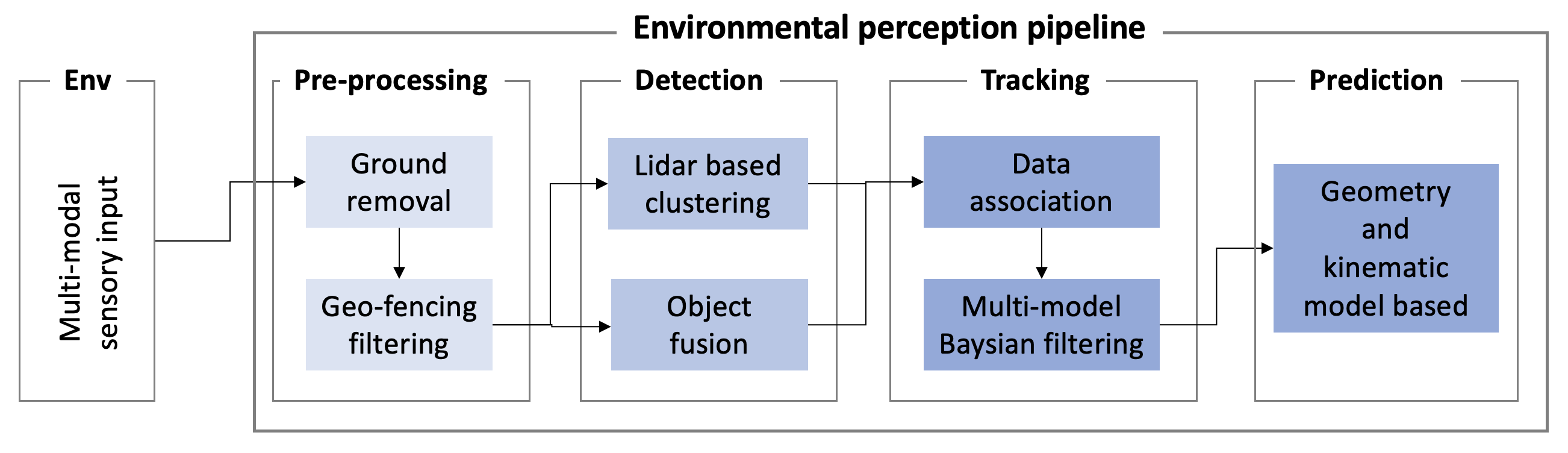}}
\caption[Schematic flow diagram for the object detection pipeline.]{Schematic flow diagram for the object detection pipeline.} \label{fig:perception_pipeline}
\end{figure}

Given the nature of high-speed autonomous driving problem, we built a multi-modal perception pipeline using the sensory system of the vehicle, as shown in Fig. \ref{fig:sensors_layout_realveh}. Our perception stack mainly comprises a GNSS-based ego state estimator and an environmental perception part, which includes detection, tracking, and prediction (see Fig. \ref{fig:perception_pipeline}). In the following sections, we will present the details of our approach and our implementation results. 


\subsubsection{Ego State Estimation}
\label{sec:localization}

Reliable state estimation is crucial for autonomous robots, especially under high-speed driving conditions. Since it is a long-studied problem, there are various ways, including Simultaneous Localization And Mapping (SLAM), odometry estimation, and GNSS-based navigation. Each approach presents advantages and weaknesses in accuracy, consistency, computational burden, etc. However, after testing SLAM-based methods \citep{shan2020lio,shan2018lego,qin2020lins} at the IMS, we concluded that SLAM-based navigation is not appropriate for the racing domain since it needs considerable computation resources. Also, it does not work well in feature-poor environments such as long straight sectors. 

\begin{wrapfigure}{r}{0.42\textwidth}
\centering
\captionsetup{justification=centering,margin=0.1cm}
\centerline{\includegraphics[width=0.315\textwidth]{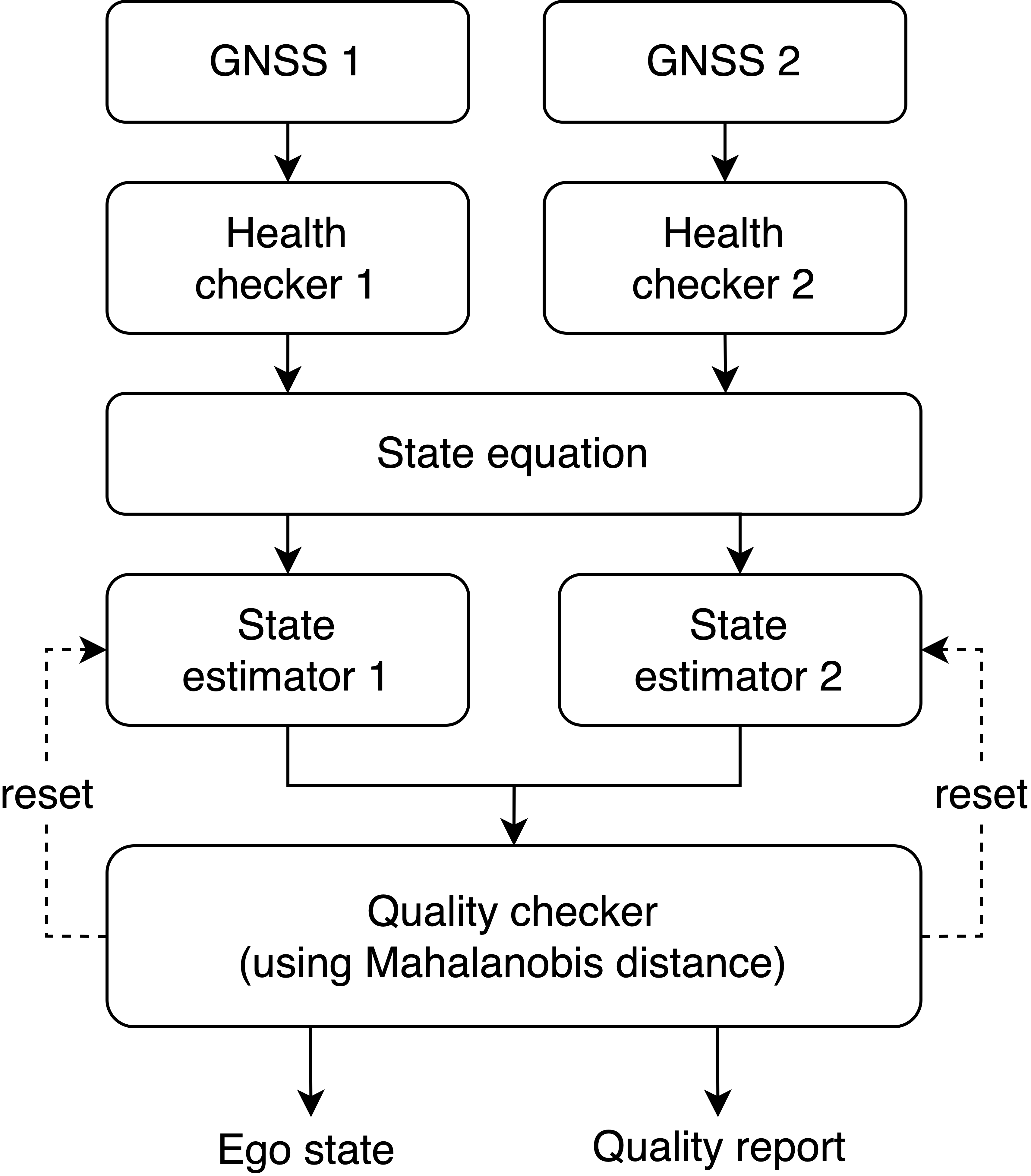}}
\caption{Resilient ego state \\estimation pipeline.}
\label{fig:localization_pipeline}
\end{wrapfigure}
Eventually, we designed our state estimator using two GNSS receivers with Kalman Filtering (KF). Our state estimator was designed to output the state information and an indicator of the solution quality for system-level resilience. The motivation behind this design choice is that, if one of the two GPS units has critical issues due to failures in signal receiving (e.g., spikes, drift, null) or hardware malfunctioning, the state-estimation task can be carried out relying on the other unit (Failure of both GPS units case should be handled differently and we will introduce our solution in Section \ref{sec:real_time_perception_control}).

Fig. \ref{fig:localization_pipeline} depicts our state estimator architecture. Two KF-based state estimation algorithms \citep{karimipour2015extended} are running in parallel using two different GNSS sources. Each algorithm estimates position, orientation, velocity, and acceleration. Two estimation outputs are passed to the quality checker. The quality checking part has two main functions related to reliable state estimation and system-level resilience: 1. measures the quality for the individual estimation outputs and selects the best. 2. reports the state estimator's health to the system status manager. We measured the quality of estimation using a Mahalanobis distance \citep{de2000mahalanobis} between estimation and sensor measurements as follows: 
\begin{equation}
\begin{aligned}
    D_h
    & \triangleq (\mathbf{x} - \mathbf{z})^T\Sigma^{-1}(\mathbf{x} - \mathbf{z})
\end{aligned}
\label{eq_mahala}
\end{equation}
where $\mathbf{x}$ is the estimation output, $\mathbf{z}$ is the sensor measurement, and $\Sigma^{-1}$ is the process covariance. We empirically set some ranges of Mahalanobis distance to represent the state estimation quality and is reported to the system status manager.
For more details of our localization module, please refer to \citep{lee2022resilient}.

\subsubsection{Opponent Detection and Tracking}
\label{sec:detection}
LiDARs and radars were used for opponent vehicle detection. Even though we had implemented a vision-based deep learning detection algorithms \citep{liu2016ssd,howard2019searching}, a few practical issues prevented us from using this method. First, we could not collect enough vision data in the real-world environment for training and validation. As an alternative, we augmented our training dataset using web-crawled data and a simulator. However, the real-world detection performance was not reliable enough. Also, external sensor fusion between cameras and LiDARs was not fast enough for high-speed driving scenarios. 
\begin{algorithm}[b!]
\caption{Height-distribution-based ground filtering algorithm}\label{alg:lidar_groud_filtering}
\begin{algorithmic}
\Require Raw LiDAR point cloud, $P$, Grid cell size, $g_c$
\Ensure Non-ground LiDAR points, $P_n$
\State $P_p \gets projectToXYplane(P)$
\State $G \gets gridDiscretization(P_p, g_c)$
\State $P_n \gets \{\}$
\For{$g \in G$}
\State $n_p \gets countNumOfPoint(g)$
\State $h_d \gets calcHeightDistribution(G,g)$
\If{$n_p > n_{thres}$ and $h_d > h_{thres}$}
    \State $R_n \gets reconstructTo3D(G,g)$
    \State $P_n = P_n \cup R_n$
\EndIf
\EndFor
\State $return P_n$
\end{algorithmic}
\end{algorithm}

\begin{figure}[t!]
\centerline{\includegraphics[width=.97\textwidth]{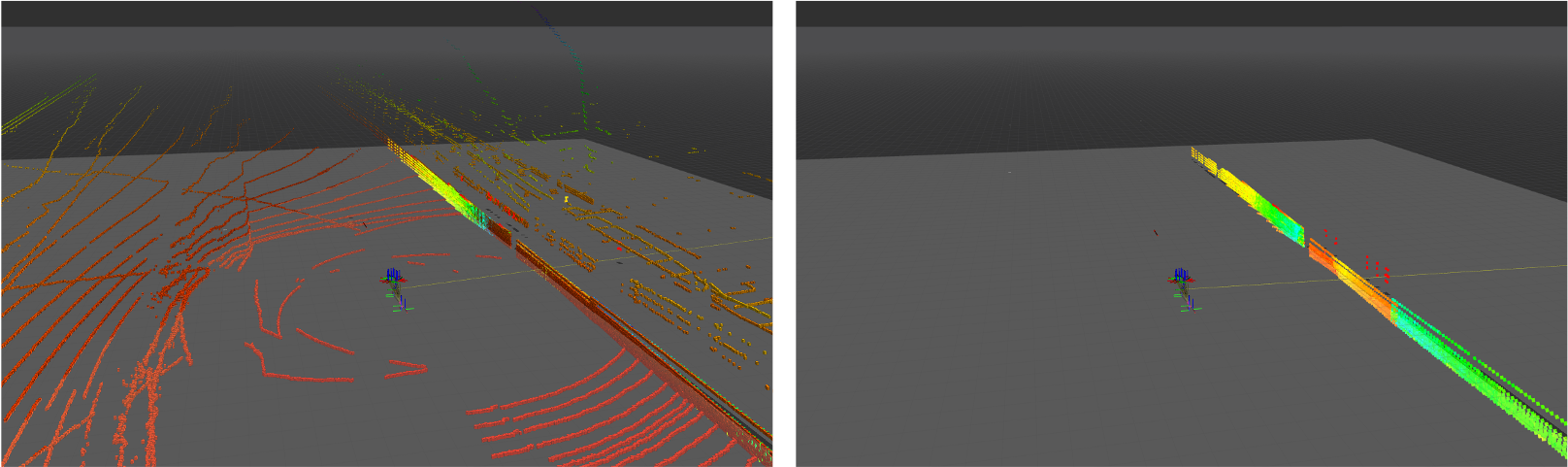}}
\caption[Ground filtering result of 3D point clouds based on height distribution. (Left) Merged all 3 LiDARs point clouds. Before ground filtering. (Right) Ground filtered result.]{Ground filtering result of 3D point clouds based on height distribution. (Left) Merged LiDARs point clouds before ground filtering. (Right) Ground filtered result.}
\label{fig:lidar_preprocessing}
\end{figure}

Our perception pipeline begins with pre-processing sensor data. The ground removal algorithm distinguishes between the ground plane and non-ground points in lidar point cloud data. The most representative process is to assume one plane within the sensing range and remove the points associated with the plane using plane fitting. However, the target track is an oval with banking in all sectors, and the change in bank angle is significant, particularly in the area entering or exiting the corners from or into the stretch. To that end, we created a non-ground point filtering algorithm based on height distribution (see Algorithm \ref{alg:lidar_groud_filtering}). 
Our pre-processing method projects three-dimensional LiDAR points into XY-plane and represents into grid cells. After that, two factors are considered: 1. Number of points in the grid, 2. Height distributions of the grid. When these two different criteria are over thresholds, we assumed that the grid is occupied by non-ground objects.
Fig. \ref{fig:lidar_preprocessing} shows the output of ground filtering on the LiDAR point cloud. In addition, a geofencing filter is applied using a track model for both LiDAR and radar data. Sensor data around 1 m from the track boundary was removed. 

After the pre-processing step, the filtered data contains only a few points corresponding to the opponent vehicle on the track. Then, we adapted a clustering algorithm that is widely used in autonomous robots for grouping and labeling points associated with an object \citep{uppada2014centroid}. The clustering of point clouds in this work is accomplished using a hierarchy-based method \citep{madhulatha2012overview}. The hierarchy-based clustering algorithm, also known as connectivity-based clustering, classifies objects based on the distance between neighboring points. However, this clustering method is sensitive to outlier points and may result in separating a single object into multiple clusters. We used two consecutive clustering steps with different connectivity criteria to obtain more accurate results. The clustering result was defined as the object's center point and maximum width and length. By comparing the finally detected width and length with the vehicle's geometry, we cross-checked the opponent's relative position on the track. 
\begin{figure}[t!]
\centerline{\includegraphics[width=0.62\textwidth]{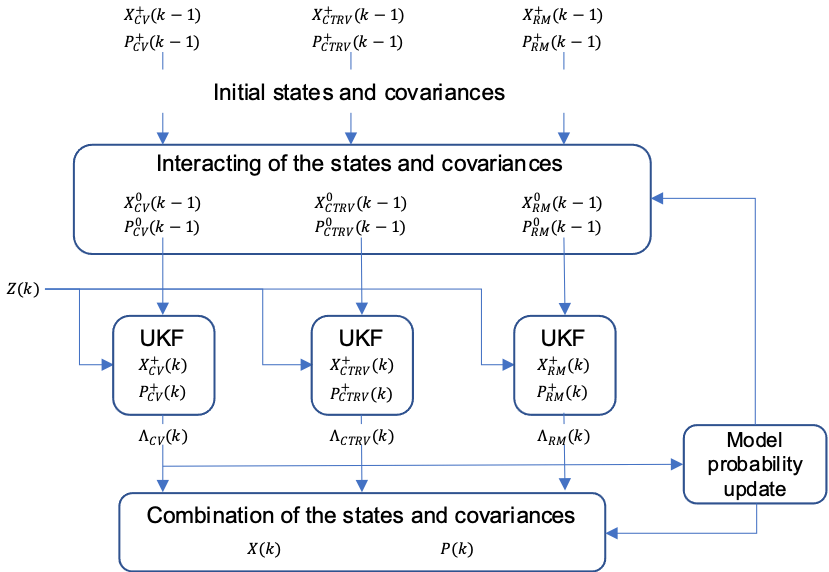}}
\caption[Schematics of one cycle of the IMM-UKF-PDAF using CV, CTRV, and RM models.]{Schematics of one cycle of the IMM-UKF-PDAF using CV, CTRV, and RM models.} \label{fig:object_tracking_diagram}
\end{figure}

Clustered objects represented in 3-Dim local coordinate are fed into the tracking module to estimate the object's dynamic states. This information is necessary for the trajectory prediction module, the last step of our environmental perception module. The tracking problem can be modeled as a filtering problem in which the object states might be noisy. Bayes filtering is one of the widely used statistical theories that can be applied directly with multiple types of models. Because of its heuristic-free approach, this filter can generally be applied to numerous robot applications \citep{thrun2002probabilistic}. Kalman Filter \citep{gutman1990tracking} is the analytical implementation of the Bayesian method that seeks to compute the optimal filter gain from its posterior density recursively. Generally, it assumes that the target objects' dynamic and posterior density at the previous observation follows Gaussian distribution, and the measurement function is linear. The Extended Kalman Filter (EKF), as well as the Unscented Kalman Filter (UKF) \citep{wan2001unscented}, are filters developed to overcome the limitation imposed by linear modeling when trying to capture object motion. 

We implemented the Interacting Multiple Model UKF Probabilistic Data Association Filter (IMM-UKF-PDAF, or IMM for short) proposed by \citep{arya20173d} in order to track robustly in varied race circumstances where $X(k)$, $P(k)$, $Z(k)$, and $\Lambda(k)$ represent state, covariance, measurement, and the likelihood for the observation at time $k$, respectively. It estimates an object's ambiguous dynamic behavior by combining several models rather than using a single motion model for the existing filter for state estimation. IMM can be made up of $n$ filters that run different models in parallel and output individual probabilities. Then, IMM uses a weighted average of each model output to calculate a single combined estimate state and its corresponding variance for the next iteration. We employ three models for object tracking: 1. Constant Velocity (CV) 2. Constant Turn Rate and Velocity (CTRV) 3. Random Motion (RM). We set the initial weights for each model to 0.5, 0.3, and 0.2, respectively. 
Fig. \ref{fig:object_tracking_diagram} schematically shows one cycle of our tracking module's operation. 

\begin{figure}[t!] 
    \begin{subfigure}{0.5\textwidth}
        \includegraphics[width=\textwidth]{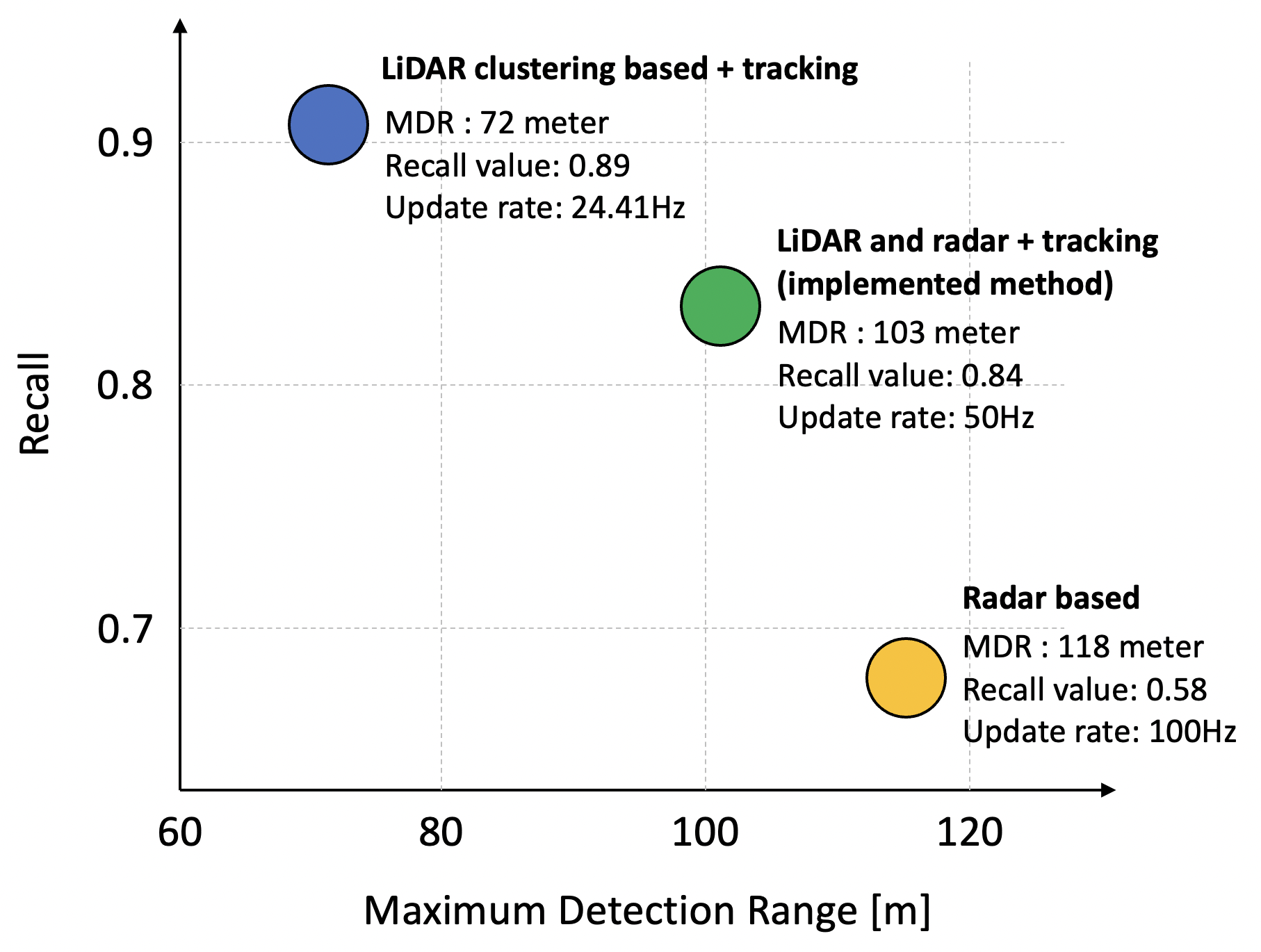}
    \end{subfigure}
    \begin{subfigure}{0.5\textwidth}
        \includegraphics[width=\textwidth]{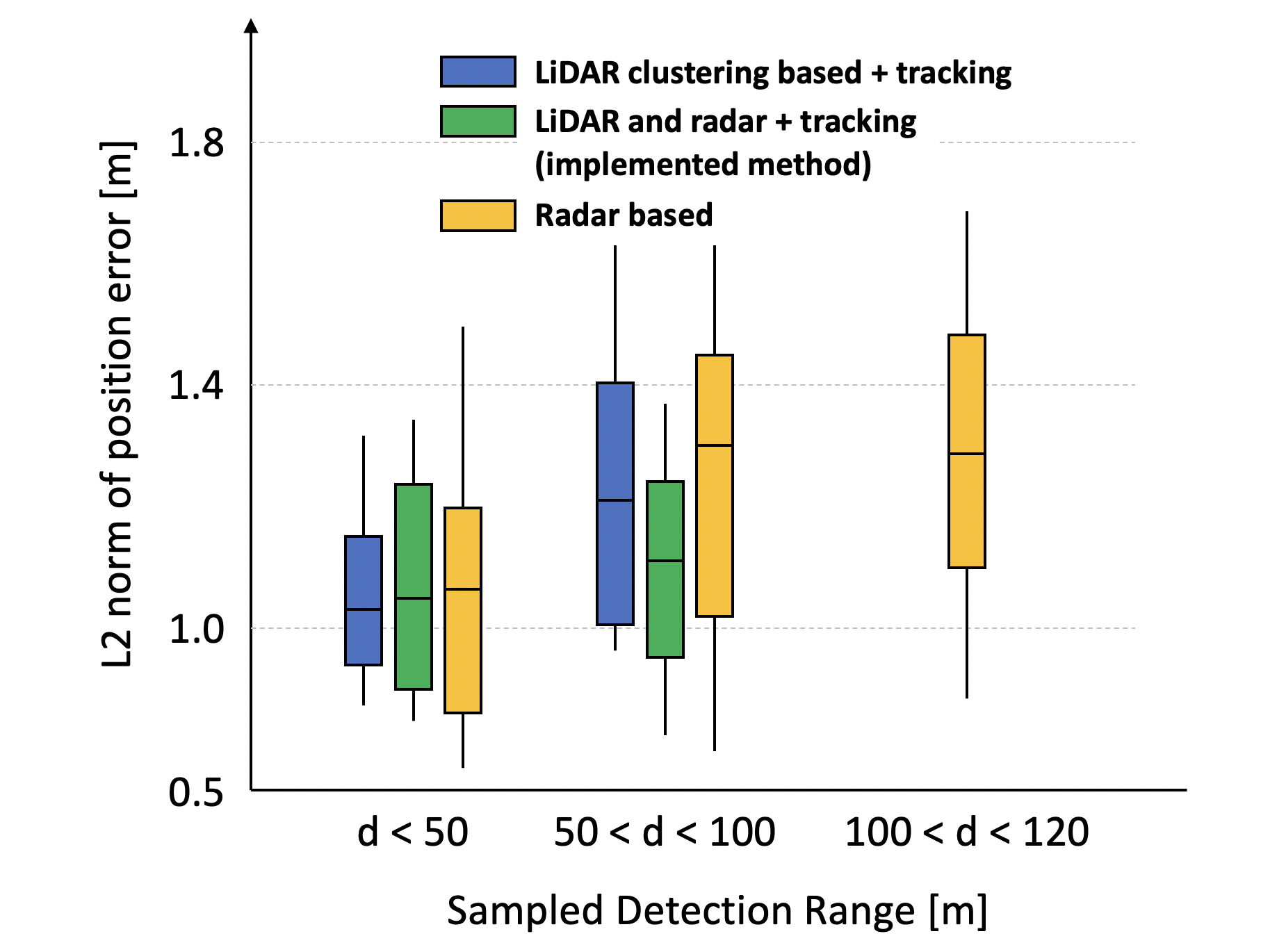}
    \end{subfigure}
\caption{Perception performance analysis results. (Left) AP and inference speed comparison results for three different methods. (Right) L2 norm of position error according to sampled detection ranges.} \label{fig:detection_comparison_result}
\end{figure}

Finally, the output of the tracking module contains the 3-Dim position and XY velocity vector in local coordinates. The results are transformed into global coordinates using the ego vehicle's status and upsampled to 50 Hz. Fig. \ref{fig:detection_comparison_result} illustrates the detection performance evaluation result during the semi-final race with \emph{PoliMOVE} (For evaluation, we used 50 samples from the logged data during the race event at LVMS. \emph{PoliMOVE} provided their GPS log, which we used as ground truth.). The LiDAR-based perception result showed the highest Average Precision (AP); however, its maximum detection range was around 70 m. On the other hand, raw radar data showed the longest detection range but lowest recall performance. Our multi-modal detection approach, as shown in the green, showed a balanced performance in terms of accuracy and range. 
Additionally, we conducted the perception performance analysis according to the range. We used the L2-norm distance to measure how our detection output is geometrically close to the ground truth. Note that, we only collected samples when the driving speed is over 100 mph. For generating the ground truth, we manually measured the position of the other vehicle using ego-vehicle's state and raw Lidar data. We divided the validation samples into three cases: closer than 50m, over 50m but less than 100m, over 100m but less than 120m. As shown in the Fig. \ref{fig:detection_comparison_result} right, our deployed method (Lidar and radar based detection and tracking) showed a well balanced in terms of detection accuracy and and its performance consistency. Given the validation case when the sampled targets are closer than 50m, Lidar only clustering based detection method showed the best performance among three implementations.

\subsubsection{Future Trajectory Prediction}
\label{sec:prediction}
Based on the perception results, autonomous vehicles must plan a safe motion trajectory considering the predicted opponent's trajectory. Accurate prediction of other vehicles' trajectories increases the safety of the autonomous vehicle and has an impact on general traffic safety, and efficiency \citep{claussmann2019review}. For these reasons, substantial research on trajectory prediction and ego-motion planning has been actively conducted in autonomous mobile robotics \citep{elbanhawi2014sampling,laumond1998robot,petti2005safe}. 

\begin{figure}[t!]
\centerline{\includegraphics[width=0.37\textwidth]{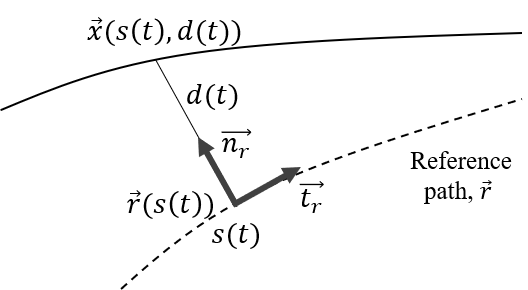}}
\caption[Representation of a trajectory in a Frenet frame.]{Representation of a trajectory in a Frenet frame.} \label{fig:frenet_origin}
\end{figure}

The majority of predictions for surrounding vehicles were made in an urban environment. However, in a racing scenario, the opponents' trajectory prediction differs from that of an urban setting. In an urban driving environment, for example, road geometry is one of the essential cues for the prediction module. In a race, however, the race line, which shows the minimum lap time, can be used for predicting an opponent's future trajectory. Furthermore, a strategic trajectory that prevents neighboring vehicles from overtaking can predict a free-racing scenario (one in which no arbitrary race restrictions apply). A few studies have modeled race as a non-cooperative game. In our previous work \citep{jung2021game}, we used the Stackelberg Game to model free-racing. We consecutively build independent Stackelberg games amongst neighboring vehicles to change the n-player game to two-player games, and each game is solved recursively. We assumed in each game that two cars were seeking to maximize their progress, with the following agent accounting for collisions. The progress term is converted to a payoff function similar to the MPC objective function. Our method was evaluated in a simulated environment and demonstrated the capability to pass in multi-agent competitive race scenarios. 

However, the real-world IAC competition was ultimately decided to be a 1:1 overtaking competition for safety reasons. Also, the race rules imposed the defending vehicle to maintain a commanded speed (by racing control) along the inner side of the track. We designed the prediction task more straightforwardly following the competition's rules, and we made the following two reasonable assumptions for prediction:
\begin{enumerate}
\item The opponent will stay on track, $\boldsymbol{X}_{track}$.
\item During the prediction, the opponent will keep lateral displacement from the track center line and velocity.
\end{enumerate}

The prediction problem is to forecast the set of future states \(P_{i=1,2,..,n} = \{ p_i^{0},p_i^{1},...,p_i^{t_{pred }} \}\) for the prediction time instant \(0\) to \(t_{pred}\) where $p_i^t = (x_i^t, y_i^t, v_i^t)$. $x, y$, and $v$ denote the spatial Cartesian coordinates x and y and velocity, respectively. We used the Frenet coordinate system for prediction which is a well-known coordinate system in trajectory planning and control theory. The Frenet frame is made up of three vectors: the normal vector $\vec{n}_r$, the tangential vector $\vec{t}_r$, and the binormal vector $\vec{b}_r$. These vectors can describe the kinematic properties of a particle moving along a continuous curve, as illustrated in Fig. $s(t)$ and $d(t)$ represents the progress alongside the reference path and the lateral displacement at time $t$, respectively. 

Assuming that the track's center line is $\vec{r_c}$, the predicted trajectory of the surrounding vehicles $\vec{x}$ is expressed in Frenet coordinate space as follows, following the aforementioned assumptions (see Fig. \ref{fig:frenet_origin}):
\begin{equation}
    \vec{x}(s(t),d(t)) = \vec{r_c}(s(t)) + d(t)\vec{n_c}(s(t))
\end{equation}
\begin{enumerate}
\item $\vec{x}(s(t),d(t)) \in \boldsymbol{X}_{track}$
\item $d(t)=d(t_0)$, $\dot{s}(t)=\dot{s}(t_0)$, $\ddot{s}(t)=0$ $\forall 0 \leq
t < t_{pred}$
\end{enumerate}

We calculated the lateral displacement from the center line of the track, $d_0 = d(t_0)$, and its progress, $s_0 = s(t_0)$. Here, we set the prediction time horizon to 4 seconds with a sampling time of 0.1 sec. Prediction results are kept even if perception results are not updated for a certain period (here, we set 2 secs) to cope with the instant loss of the tracked objects. To this end, the internal loop updates the trajectory by propagating the object's status using the previous prediction result. Algorithm \ref{alg:euclid} shows the principal steps of our prediction process.

\begin{algorithm}[h!]
\caption{Prediction solution updating algorithm}\label{alg:euclid}
\begin{algorithmic}[1]
\Procedure{$updatePrediction$}{}
\While{True}
\State $t_p \gets GetCurrentTime()$ \Comment{Update time.}
\State $F_p \gets CheckPerceptionOutput()$ \Comment{Check whether there is perception output.}
\If{$F_p$} 
    \State $P_c \gets GetPerceptionOutput()$ \Comment{This only activate when there is perception result.}
    \State $t_c \gets GetCurrentTime()$ \Comment{Update time when the perception result is updated.}
    \State $P_p \gets GenPredictionOutput(P_c)$
    \State $return(P_p)$
\Else
\If{$P_p \neq \emptyset \text{ and } t_c + t_{thres} \leq t_p$}
\State $P_p \gets PropagateOnPreviousResult(P_p)$
\State $return(P_p)$
\Else
\State $P_p \gets Clear()$
\EndIf
\EndIf
\EndWhile \label{alg:predictionUpdate}
\EndProcedure
\end{algorithmic}
\end{algorithm}

\subsection{High-speed Overtaking Planner}

\subsubsection{Global Race Line Generation}
\label{sec:raceline}
The definition of the race line is the path that minimizes the lap time considering the vehicle dynamics and track geometry. Following this definition, the race line generation problem can be modeled as a nonlinear optimization problem as follows:
\begin{equation}
\label{eq:raceline_obj}
    \begin{aligned}
    \text{Minimize } & t = \int dt = \int \frac{dt}{ds}ds = \int \frac{1}{v} ds \\
    \text{subject to\quad} &\kappa v^2 - \mu g \leq 0 \\
     &\kappa \leq \kappa_{max} \\
     &v \leq v_{max} \\
     &a_{min} \leq a \leq a_{max} \\
    \end{aligned}
\end{equation}
where $t$ is the total time, $s$ is the vehicle's travel length, $v$ is the vehicle's velocity, $a$ is the acceleration, $\mu$ is the tire-road friction coefficient, and $\kappa$ denotes the curvature. 

Various approaches to calculating the optimal race line have been proposed \citep{christ2021time,dal2019comparison,lovato2022three}. Most of the previous researches utilized nonlinear vehicle and tire dynamic models and solved the Eq. \ref{eq:raceline_obj} directly via optimization framework, and the properties of the vehicle were incorporated into the constraint functions. However, acquiring the model parameters is not trivial. Also, the performance degradation due to model mismatch is one of the well-known limitations of "model-based" approaches. \citep{heilmeier2019minimum} demonstrated experimentally that the minimum curvature trajectory performs nearly as well as the model-based optimal racing line. The path's curvature is minimized, which maximizes the achievable velocity. 

\begin{figure}[t!]
\centerline{\includegraphics[width=0.63\textwidth]{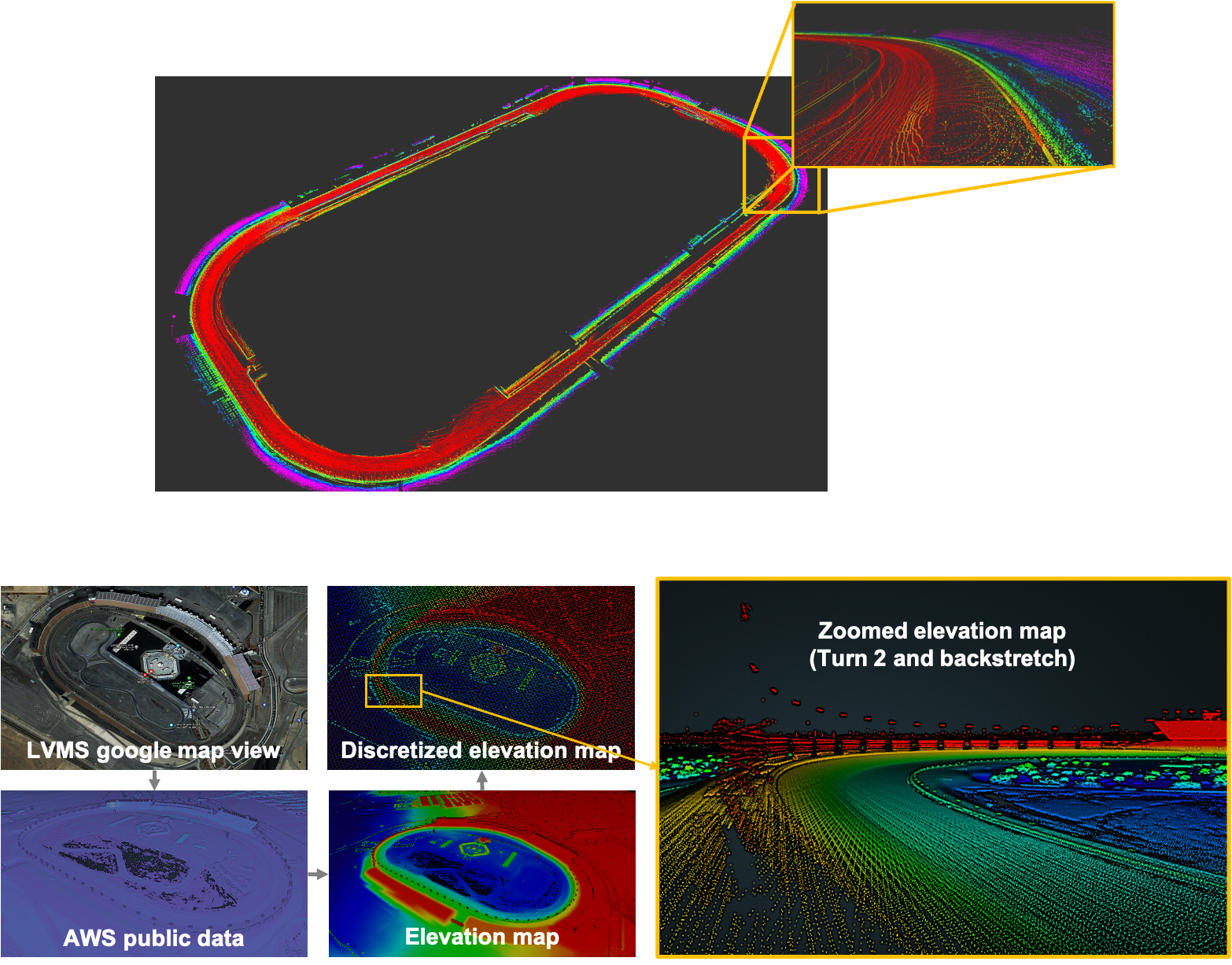}}
\caption[Visualization of track modeling results. (Top row) Indianapolis Motor Speedway(IMS) mapping result. (Bottom row) Las Vegas Motor Speedway(LVMS) track modeling flow using USGS data.]{Visualization of track modeling results. (Top row) Indianapolis Motor Speedway(IMS) mapping result. (Bottom row) Las Vegas Motor Speedway(LVMS) track modeling flow using USGS data.} \label{fig:mapmap} 
\end{figure}

We generated the race line using the minimum curvature trajectory following the idea from \citep{heilmeier2019minimum}. On top of that, we also considered the travel distance since the lap time is a function of speed and distance. To this end, we started from collecting the track boundaries. We modeled the track using LEGO LOAM \citep{shan2018lego} and the United States Geological Survey (USGS) LiDAR point cloud data. We manually extracted the track's inner and outer boundary points from the 3D map and interpolated them using a cubic spline model. Fig. \ref{fig:mapmap} shows our map-building results and examples of USGS data. After that, we generated the geometrically centered line of the track, the reference line representing the track, using the average operation on the sampled two-track boundary point sets, and it can be written as follows:
\begin{equation}
    \vec{r}_{n} = \vec{p}_{n} + \alpha_n \vec{n}_{n}
\end{equation}
where $\vec{p}_{n}$ is the center line point, $\vec{n}_{n}$ is the unit normal vector, and $\alpha_n$ is the distance to the track boundary.

\begin{table}[t!]
\centering
\caption[Results of estimated lap times under different weights, $\zeta$, and maximum speed limit conditions. If the maximum speed of the autonomous system is below the dynamics limit of the vehicle itself, it can be seen that the minimum curve-based race line does not always show the minimum lap time.]{Results of estimated lap times under different weights, $\zeta$, and maximum speed limit conditions. If the maximum speed of the autonomous system is below the dynamics limit of the vehicle itself, it can be seen that the minimum curve-based race line does not always show the minimum lap time.}
\label{tab:zeta_laptime}
\begin{tabular}{cccccc}
\hline
\multirow{2}{*}{Track} & \multirow{2}{*}{\begin{tabular}[c]{@{}c@{}}Maximum velocity\\ constraint\end{tabular}} & \multicolumn{4}{c}{Estimated lap time [sec]}\\ \cline{3-6}& & \multicolumn{1}{c}{$\zeta$=0.0} & \multicolumn{1}{c}{$\zeta$=0.3} & \multicolumn{1}{c}{$\zeta$=0.6} & $\zeta$=0.8 \\ \hline \hline
\multirow{2}{*}{LVMS}  & 220 ${km}/{h}$ & \multicolumn{1}{c}{44.003}  &\multicolumn{1}{c}{43.197}     & \multicolumn{1}{c}{\textbf{42.847}}    &   42.889   \\ \cline{2-6}
& 170 ${km}/{h}$ & \multicolumn{1}{c}{56.266}  & \multicolumn{1}{c}{\textbf{56.149}}     & \multicolumn{1}{c}{56.423}    & 56.501     \\ \hline
\end{tabular}
\end{table}

Given the track model, minimum curvature trajectory was generated using open source implementation from \citep{heilmeier2019minimum}. Furthermore, the shortest path was incorporated using a simple geometric-weighted sum operation. Assuming that the minimum curvature race line is $P_{mc}$ and the shortest travel distance path is $P_{sd}$, our final race line can be represented as follows:
\begin{equation}
    P_c = \zeta P_{mc} + (1-\zeta)P_{sd} 
\end{equation}
where $\zeta$ is the geometric weight parameter between two different paths. Note that, since our target track is an oval track, we set the inner boundary as the shortest path. 

We verified our idea by experimentally measuring the lap time of $P_c$ generated under various conditions using a high-fidelity racing simulator. Table \ref{tab:zeta_laptime} shows the estimated lap time results. Note that estimated lap times can be different from the real-world testing results. However, we can see that incorporating the shortest path into the race line can help to reduce the lap time in certain operation conditions (e.g., maximum speed).

\subsubsection{Local Trajectory Planning for High-speed Overtaking}
\label{sec:local_planning}
Overtaking is referred to as the core of racing. Professional race car drivers simultaneously push the vehicle to the limit while adhering to the race rules and performing strategic moves based on contextual understandings of opponents' intentions. The local trajectory planner takes responsibility for high-speed overtaking in our autonomy stack. 

\begin{figure}[t!]
\centerline{\includegraphics[width=0.6\textwidth]{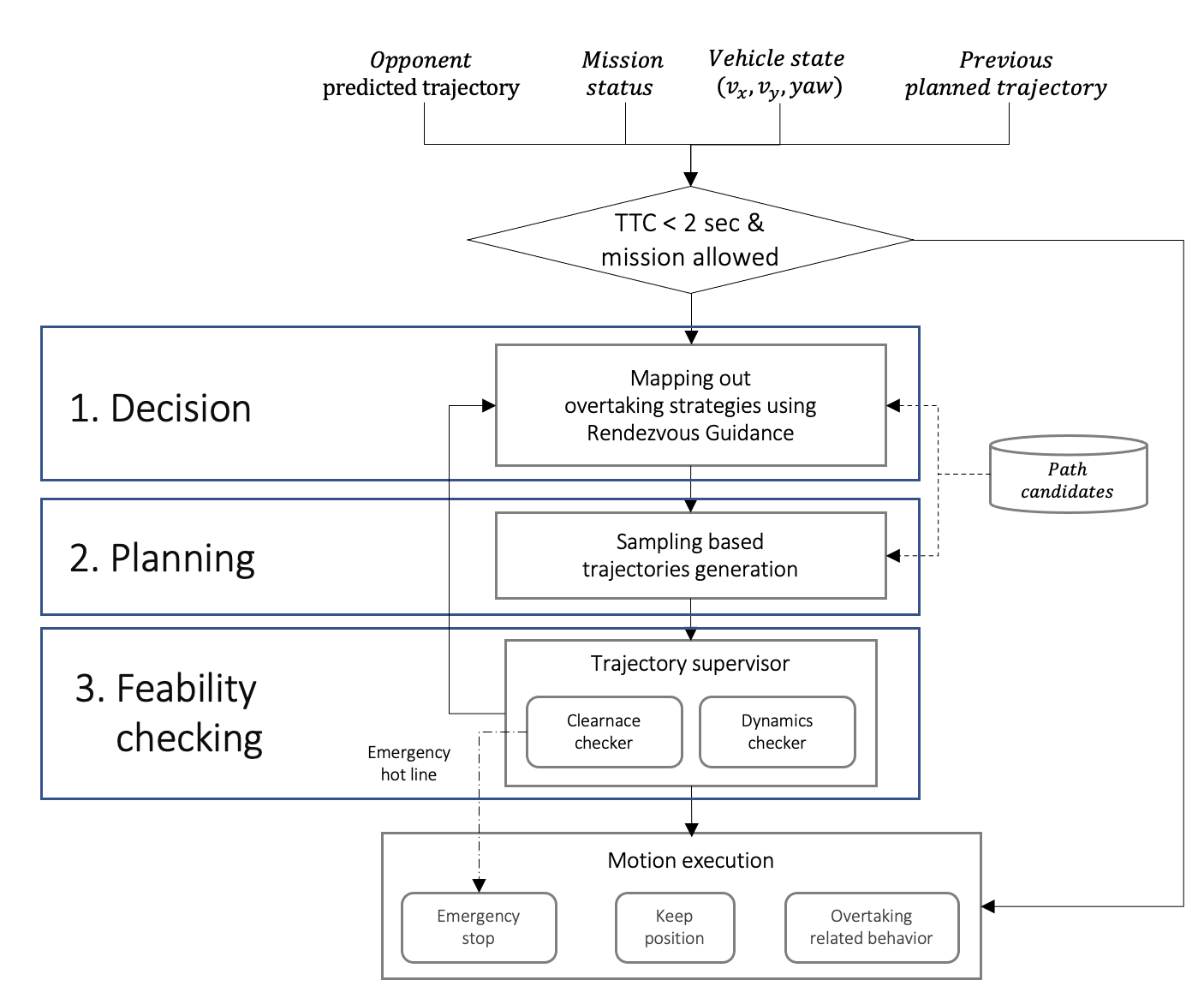}}
\caption[Schematic diagram of hierarchical overtaking trajectory planning module.]{Schematic diagram of hierarchical overtaking trajectory planning module.} \label{fig:overtaking_overview}
\end{figure} 

The proposed local trajectory planner is primarily made up of three hierarchical module stages: 1. decision, 2. planning, 3. feasibility checking. The predicted trajectory of the opponent, the current ego states and race flag (provided from the race control), and the planned trajectory from the previous step were used as inputs. Our planner starts by making overtaking decisions based on the current ego state and the predicted opponent's trajectory. The decision to attempt overtaking or not is crucial in high-speed driving scenarios since it can not be accomplished even though it is collision-free. To decide whether overtaking can be safely done under our operation setup and system capability, we adapted the Rendezvous Guidance (RG) \citep{kunwar2006motion} and applied it to the overtaking problem. The decision module determines whether overtaking is possible within a given time frame. Then, the decision outcome is passed to the sampling-based trajectory generation module. Our trajectory generation module plans the optimal jerkiness trajectory \citep{werling2010optimal} considering the driving speed and the vehicle dynamics. As the last step, the collision and tire model-based feasibility were checked before execution. Fig. \ref{fig:overtaking_overview} depicts the overall structure of our hierarchical planning module. In the following, we will deliver details of each step.  

\begin{wrapfigure}{r}{0.5\textwidth}
\centering
\captionsetup{justification=centering,margin=0.5cm}
\centerline{\includegraphics[width=0.43\textwidth]{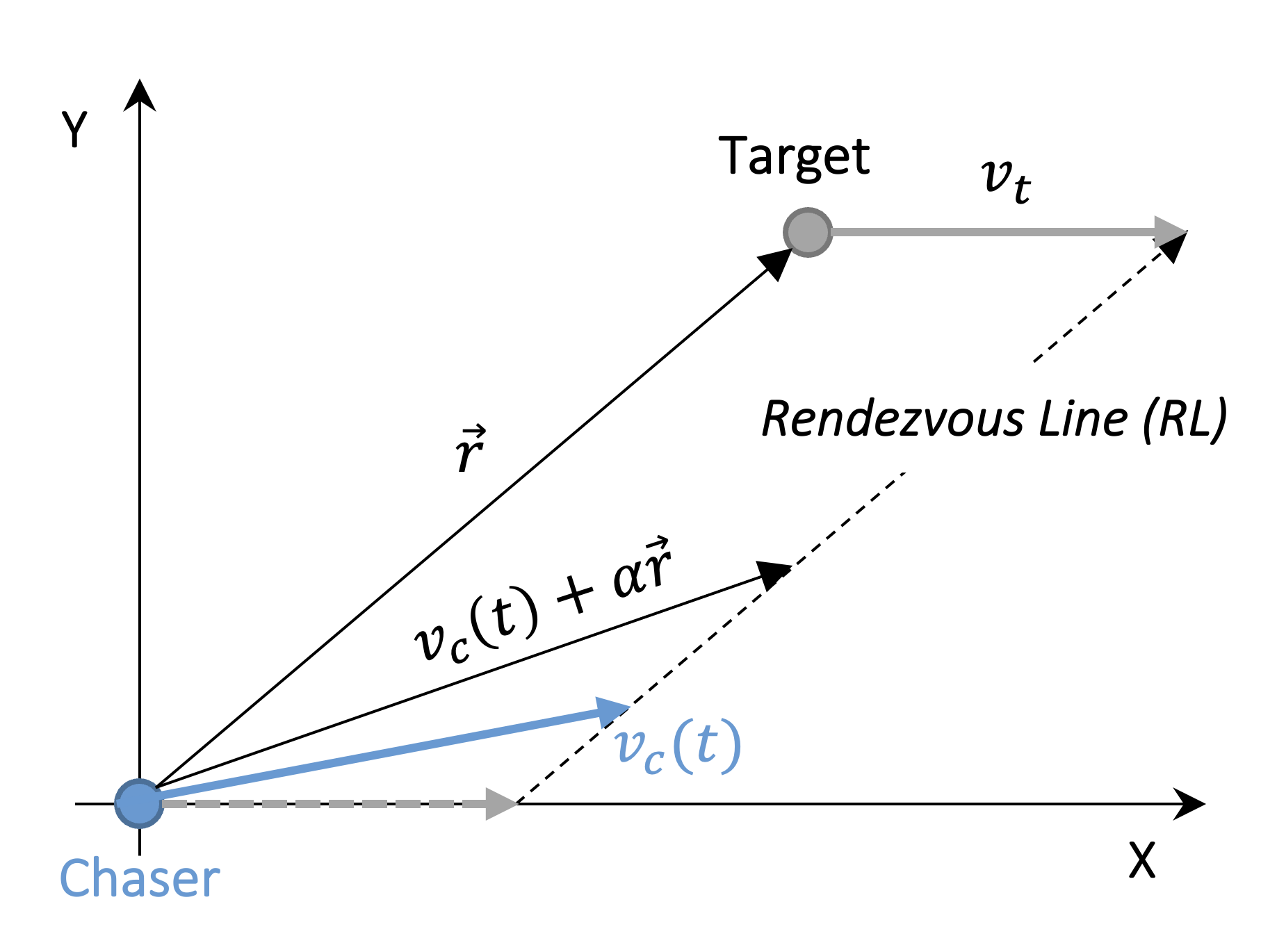}}
\caption{Construction of rendezvous line \\based on parallel-navigation law.}
\label{fig:rendezvous_line}
\end{wrapfigure}
\paragraph{Rendezvous Guidance based Overtaking Decision Making}
The proposed overtaking decision-making algorithm is inspired by \citep{ghumman2008guidance} and is based on the RG \citep{kunwar2006motion}. The rendezvous problem is well-known as a chase-target problem in which the target does not require evading acceleration due to common space debris. RG was used for the space docking mission. Analytically, it has been demonstrated that RG provides the optimal solution for rendezvous with immobile targets based on the parallel navigation law. There are two types of participants in an autonomous rendezvous: chaser and target (See Fig. \ref{fig:rendezvous_line}). Given current chaser's velocity, $v_c$, and target vehicle's velocity, $v_t$, RG is responsible for generating the velocity control signal of the chaser that will eventually match these two plants. According to the parallel-navigation law, the relative velocity between the chaser and the target, $\vec{\dot{r}}$, should remain parallel to the Line of Sight (LOS), $\vec{r}$. If this condition is satisfied, the chaser's distance from the target will decrease until they collide.

The parallel-navigation law is defined by the equations below. Eq. \ref{eq:pn1} guarantees that $\vec{r}$ and $\vec{\dot{r}}$ remain colinear, whereas Eq. \ref{eq:pn2} guarantees that the chaser does not recede from the chaser. Both Equations can be solved for $\vec{\dot{r}}$ in a parametric form using Eq. \ref{eq:pn3} where $\alpha$ is a positive real number.
The RG's output is a time-optimal velocity command for the chaser under the parallel-navigation law. As illustrated in Fig. \ref{fig:rendezvous_line}, if the chaser maintains a velocity command parallel to the Rendezvous Line(RL), the direction of LOS remains constant, ensuring positional matching between the chaser and the target.
\begin{equation}
\label{eq:pn1}
\vec{r}\times\vec{\dot{r}} = 0
\end{equation}
\begin{equation}
\label{eq:pn2}
\vec{r}\cdot\vec{\dot{r}} < 0
\end{equation}
\begin{equation}
\label{eq:pn3}
\vec{\dot{r}}= -\alpha\vec{r}
\end{equation}

\begin{figure}[t!]
\centerline{\includegraphics[width=.7\textwidth]{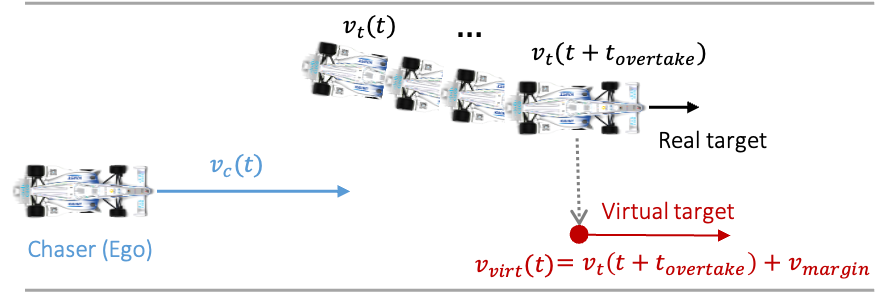}}
\caption[Conceptual visualization of virtual target generation for overtaking decision making.]{Conceptual visualization of virtual target generation for overtaking decision making.} \label{fig:virtual_target}
\end{figure}

\begin{wrapfigure}{r}{0.6\textwidth}
\centering
\captionsetup{justification=centering,margin=0.5cm}
\centerline{\includegraphics[width=0.58\textwidth]{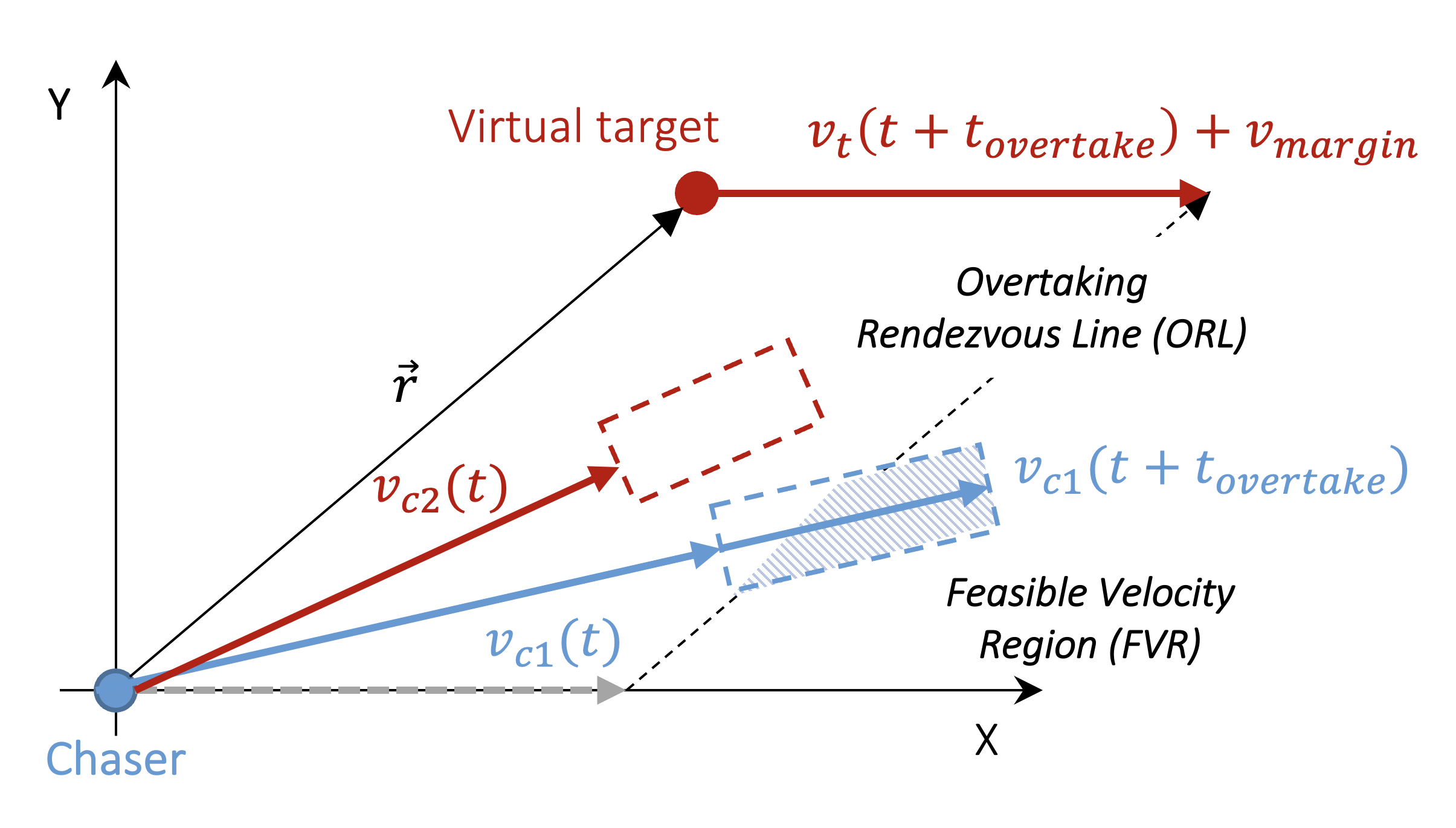}}
\caption[Construction of ORL and FVR.]{Construction of ORL and FVR.} \label{fig:ORL_FVR}
\end{wrapfigure}

The RG, position matching guidance, is applied to an overtaking problem by creating a virtual target (See Fig. \ref{fig:virtual_target}). Our virtual target is spawned parallel to the predicted trajectory at $t_{overtake}$ point (here, we set 6 seconds). 
Using the virtual target, we modified the RG to determine whether our autonomy is capable of overtaking or not. Fig. \ref{fig:ORL_FVR} visualized the modified RG plot. 

Considering the maximum acceleration based on the vehicle's current speed, race flag, and track sector, the Feasible Velocity Region (FVR) is generated. Here, we assumed that the vehicle can accelerate up to 80\% of the maximum acceleration for generating the FVR as a safety-performance balance. 
Finally, the decision module outputs the overtaking trigger signal if the FVR is beyond the Overtaking Rendezvous Line (ORL), which indicates that overtaking is roughly feasible under the current operating configuration. 
As examples, we visualized two different chaser's velocity vectors in Fig. \ref{fig:ORL_FVR} using $v_{c1}$ and $v_{c2}$. With the $v_{c1}$, FVR with blue boxed area is over the ORL, which can be interpret that our vehicle can be proceed more than the virtual target position at $t_{overtake}$. On the other hand, $v_{c2}$ with the red boxed area is not long enough to cross the ORL, which means that the ego vehicle will be still behind during the overtaking horizon. As we briefly mentioned before, the size of FVR is decided by multiple factors including vehicle states, race flag, and track sectors. Note that these multiple factors were chosen heuristically. 

\paragraph{Minimum Jerkiness Overtaking Trajectory Generation}
In this step, the ego-motion trajectory, which will be driven in the near future, is generated. We take the overtaking maneuver as the path switching of the race line and one of the path candidates. Five different reference paths parallel to the track's center line, $P_{candidate} \in P_ {left},P_{left,center},P_{center}, P_{right,center},P_{right}$ were used. Our local trajectory planner is in charge of generating the connecting trajectory segment, $\tau = \{p_{t_0}, p_{t_1},...,p_{t_{ot}}\}$, between the current and the target path, where $p_{t}$ is the set of position at time $t$. 

Our planner aims to generate a minimum jerkiness trajectory to safely and quickly merge to the target path. Jerkiness is a widely used indicator of comfort in autonomous passenger vehicles, but in the case of a race vehicle, it also impacts the vehicle's stability. Inspired by \citep{werling2010optimal}, we generated the minimum jerkiness trajectory by using lateral and longitudinal models in Frenet-frame coordinates as follows:
\begin{equation}
\begin{aligned}
 d(t) &= c_{d0} + c_{d1}t + c_{d2}t^2 + c_{d3}t^3 + c_{d4}t^4 + c_{d5}t^5, \\
 s(t) &= c_{s0} + c_{s1}t + c_{s2}t^2 + c_{s3}t^3 + c_{s4}t^4
\end{aligned}
\end{equation}
where $c_{i,i=\{d0,d1,d2,d3,d4,d5\}}$ and $c_{i,i=\{s0,s1,s2,s3,s4\}}$ are the coefficients of the lateral and longitudinal components, respectively (we refer to \citep{werling2010optimal} for the theoretical proof). Since our goal for the local planner is to merge to the target path, we can set the end condition as $d(t_{ot})=0$. Using a set of time, $t_{ot} \in {6,4,2}$, we generated trajectories with a variety of longitudinal patterns. For the sake of simplicity, we assumed acceleration is zero during planning. By solving Equations \ref{eq:5th} and \ref{eq:4th}, lateral and longitudinal polynomial parameters can be easily calculated given the initial and final states.
\begin{equation}
\label{eq:5th}
\begin{bmatrix}
t_0^5   & t_0^4 & t_0^3 & t_0^2 & t_0^1 & 1 \\ 
t_{ot}^5   & t_{ot}^4 & t_{ot}^3 & t_{ot}^2 & t_{ot}^1 & 1 \\ 
5t_0^4  & 4t_0^3 & 3t_0^2 & 2t_0^1 & 1 & 0 \\ 
5t_{ot}^4  & 4t_{ot}^3 & 3t_{ot}^2 & 2t_{ot}^1 & 1 & 0 \\
20t_0^3 & 12t_0^2 & 6t_0^1 & 2 & 0 & 0 \\ 
20t_{ot}^3 & 12t_{ot}^2 & 6t_{ot}^1 & 2 & 0 & 0 \\ 
\end{bmatrix}
\cdot
\begin{bmatrix}
 c_{d5} \\  c_{d4} \\  c_{d3} \\  c_{d2} \\ c_{d1} \\  c_{d0} \\ 
\end{bmatrix}
=
\begin{bmatrix} 
d_0 \\  d_{t_{ot}}=0 \\  \dot{d}_0 \\  \dot{d}_{t_{ot}}=0 \\ \ddot{d}_0 \\  \ddot{d}_{t_{ot}}=0 \\ \end{bmatrix}
\end{equation}
\begin{equation}
\label{eq:4th}
\begin{bmatrix}
t_0^4   & t_0^3 & t_0^2 & t_0^1 & 1 \\ 
4t_0^3  & 3t_0^2 & 2t_0^1 & 1 & 0 \\ 
4t_{ot}^3  & 3t_{ot}^2 & 2t_{ot}^1 & 1 & 0 \\
12t_0^2 & 6t_0^1 & 2 & 0 & 0 \\ 
12t_{ot}^2 & 6t_{ot}^1 & 2 & 0 & 0 \\ 
\end{bmatrix}
\cdot
\begin{bmatrix}
  c_{s4} \\  c_{s3} \\  c_{s2} \\ c_{s1} \\ c_{s0} \\ 
\end{bmatrix}
=
\begin{bmatrix}
  s_0 \\  \dot{s}_0 \\  \dot{s}_1=\dot{s}_0 \\ \ddot{s}_0 \\  \ddot{s}_1=0 \\
\end{bmatrix}
\end{equation}

\begin{algorithm}[b!]
\caption{Trajectory clearance checking}\label{alg:sat}
\begin{algorithmic}[1]
\State $C \gets False$ \Comment{Clear collision flag.}
\For{$P_e \in Ego_{plan}$}
\For{$P_o \in Oppo_{pred}$}
\State $F \gets SAT(P_e,P_o)$
\If{$F == True$} \Comment{If any intersection exist.}
\State $t(P_e,P_o) \gets calcTimeDiff(t(P_e,P_o))$ 
\If{$t(P_e,P_o) < t_{thres}$} \Comment{Check time difference, here, we set $t_{thres}$ as 1 sec.}
\State $return \quad C \gets True$
\EndIf
\EndIf
\EndFor
\EndFor
\State $return \quad C$
\end{algorithmic}
\end{algorithm}

\begin{figure}[t!]
\centerline{\includegraphics[width=.7\textwidth]{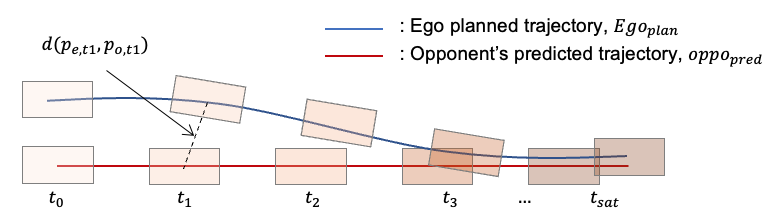}}
\caption[Collision detection based on opponent's predicted trajectory using SAT algorithm.]{Collision detection based on opponent's predicted trajectory using SAT algorithm.} \label{fig:sat}
\end{figure}

\paragraph{Trajectory Feasibility Checker}
The last step of the proposed overtaking planner verifies whether the generated trajectory collides with any predicted trajectory and also is dynamically feasible. For collision checking, a Separating Axis Theorem (SAT) was used \citep{gottschalk1996separating}. Fig. \ref{fig:sat} and the Algorithm \ref{alg:sat} illustrate the collision checking procedure visually and algorithmically. 
For each point in both generated and predicted trajectories, a box shape polygon was created based on the vehicle dimension (See Fig. \ref{fig:sat}), and collisions were detected by checking the intersection between two polygons using SAT. When trajectory clearance checking module confirm that there is no collision, the planned trajectory is investigated whether it also satisfies the tire constraint using the centripetal force equilibrium \citep{pacejka2005tire}. 
Finally, a collision-free and dynamically feasible trajectory is input to the controller to generate control commands to follow accurately.

\subsection{Resilient Control Stack}
\subsubsection{Overview}
\begin{figure}[t!]
\centerline{\includegraphics[width=.85\textwidth]{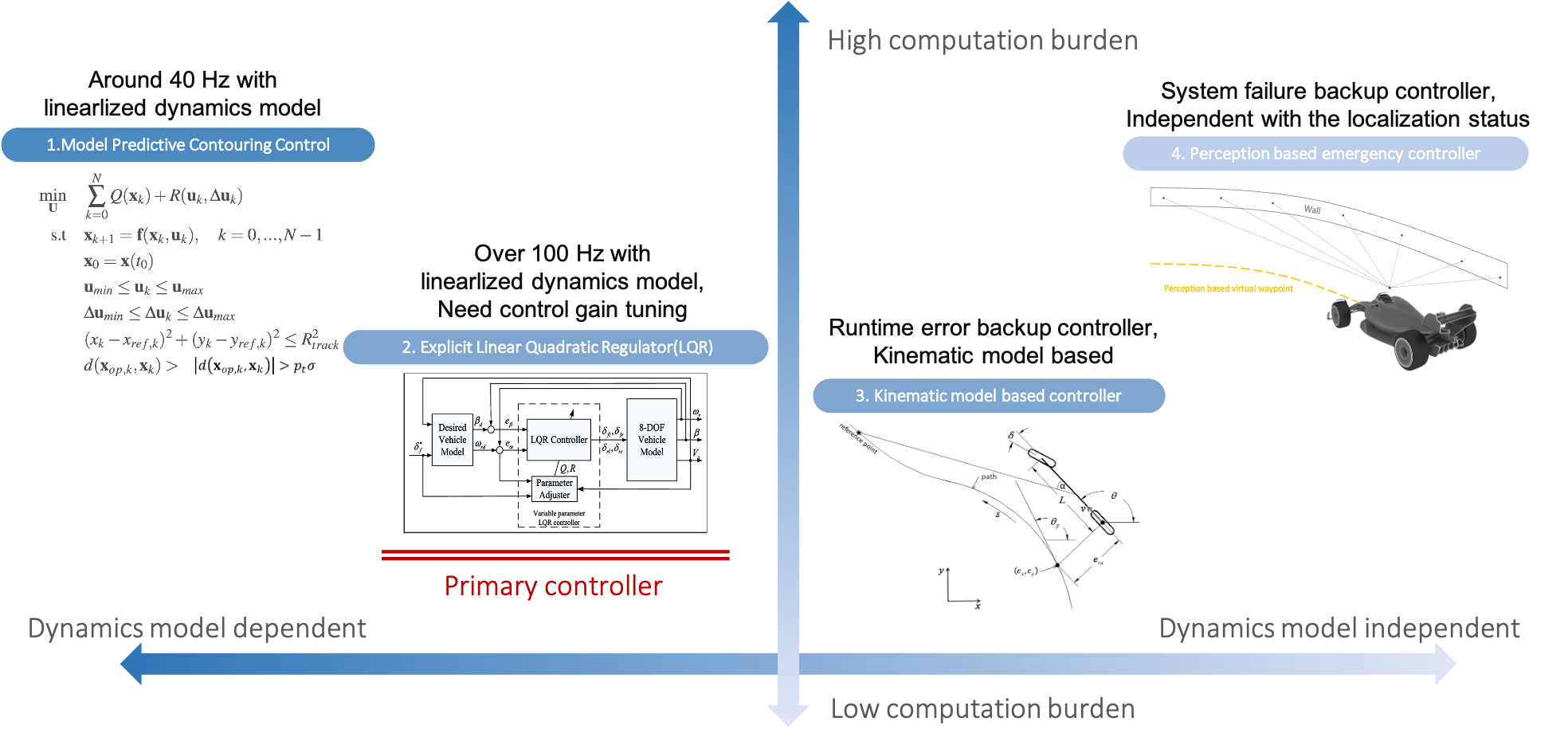}}
\caption[Overview of the control stack, including brief descriptions of each controller. Each controller is assigned a position based on the level of computational burden and model usage. Regarding real-time computability and performance, the LQR-based longitudinal and lateral controller was chosen as the primary controller.]{Overview of the control stack, including brief descriptions of each controller. Each controller is assigned a position based on the level of computational burden and model usage. In terms of real-time computability and performance, the LQR-based longitudinal and lateral controller was chosen as the primary controller.} \label{fig:control_arch}
\end{figure}

The control stack is in charge of generating throttle/brake position and steer angle to follow the planned trajectory accurately. Our high-level control stack is built on various control algorithms for overall system resilience. Control command was calculated strictly at 100 Hz and passed to the drive-by-wire system. Fig. \ref{fig:control_arch} visualizes the various types of high-level controllers and their configurations within our control stack. 

We noted that MPCC was integrated but only for the simulation race and is not be covered in this paper. For more details of our MPCC controller, refer to \citep{liniger2015optimization,jung2021game}. We used a linear–quadratic regulator controller (LQR) as a primary controller for the real-world races. To ensure real-time operability, we created the state feedback gain as a look-up table during the offline optimization. The proposed stack also includes the Stanley controller \citep{thrun2006stanley} and the pure pursuit controller \citep{coulter1992implementation}. These controllers are based on a simple kinematic model, have a simple implementation, and are exceptionally computationally light. These controllers are used as a backup solution when the primary controller has any issues (e.g., algorithm crash, calculation delay). The solely perception-based emergency controller is another resilience feature of our control stack. As the name suggested, it is independent of the global localization quality and only uses the locally sensed data (here, we used LiDARs points). It is designed to activate when the system status manager reports a fatal error or bad quality from the localization module.

Calculated steer angle and desired acceleration from the high-level controller are passed to the low-level controller to actuate throttle, brake, and steer. Dallara-AV21 is equipped with Schaeffler Paravan SpaceDrive, Drive-by-Wire (DBW) system. For the lateral control, it takes the desired steer angle for the position control. The acceleration command was converted to the pedal positions using engine torque and gear map for the longitudinal control. Implementation details are provided in the following sections.

\subsubsection{Explicit LQR-based Controller}
As the primary controller of the control stack, we used the LQR-based full-state-feedback controller using linearized system models \citep{lewis2012optimal,spisak2022robust,rajamani2011vehicle}.

For the lateral motion model, we used the lateral dynamic bicycle model, whose state elements are represented in terms of position and orientation errors with respect to a reference path. The position error, $e_y$, and the heading error, $e_{\psi}$, are modeled as follows:
\begin{equation}
\begin{aligned}
    \frac{d}{dt}
    \begin{bmatrix} 
       e_{y} \\ \dot{e}_{y} \\ e_{\psi} \\ \dot{e}_{\psi}
    \end{bmatrix} &= 
    \begin{bmatrix} 
     0 & 1 & 0 & 0 \\
     0 & -\frac{2 C_{\alpha f} + 2 C_{\alpha r}}{m V_x} & \frac{2 C_{\alpha f} + 2 C_{\alpha r}}{m} & -\frac{2 C_{\alpha f} l_f - 2 C_{\alpha r} l_r}{m V_x} \\
     0 & 0 & 0 & 1 \\
     0 & -\frac{2 C_{\alpha f} l_f - 2 C_{\alpha r} l_r}{I_z V_x} & \frac{2 C_{\alpha f} - 2 C_{\alpha r}}{I_z} & -\frac{2 C_{\alpha f} l_f^2 + 2 C_{\alpha r} l_r^2}{I_z V_x}
    \end{bmatrix}
    \begin{bmatrix} 
       e_{y} \\ \dot{e}_{y} \\ e_{\psi} \\ \dot{e}_{\psi}
    \end{bmatrix} \\
    &+
    \begin{bmatrix} 
        0 \\ \frac{2 C_{\alpha f}}{m} \\ 0 \\ \frac{2 C_{\alpha l_f}}{I_z}
    \end{bmatrix} \delta + 
    \begin{bmatrix} 
    0 \\ -\frac{2 C_{\alpha f} l_f - 2 C_{\alpha r} l_r}{m V_x} - V_x \\ 0 \\ -\frac{2 C_{\alpha f} l_f^2 - 2 C_{\alpha r} l_r^2}{I_z V_x} 
    \end{bmatrix} \dot{\psi}_{des} + 
    \begin{bmatrix} 
    0 \\ g \\ 0 \\ 0 
    \end{bmatrix} sin(\phi)
\end{aligned}
\label{eq:lateral_model}
\end{equation}
where $C_{\alpha f}, C_{\alpha r}$ are the cornering stiffness of the front and rear tires. The model consists of differential equations with respect to the state vector $[e_{y}, \dot{e}_{y}, e_{\psi}, \dot{e}_{\psi}]^T$ and steering angle control $\delta$. The other terms are governed by the desired yaw rate $\dot{\psi}_{des}$ and bank angle of the track $\phi$, which are given by the reference path and track condition. As $\dot{\psi}_{des}$ and $\phi$ are not included in the state vector, we compensated those terms by using a feedforward control component to minimize the steady-state error \citep{rajamani2011vehicle}.
Without considering the compensated terms, since the mass $m$, yaw moment of inertia $I_z$, and distance from the center of gravity to the front and rear axle $l_f, l_r$ are constants, the model can be well-formed as a state space model, assuming the current longitudinal velocity $V_x$ is constant.

Our LQR-based controller's state space model can be then represented as follows:
\begin{equation}
\begin{aligned}
   \dot{x_t} = A x_t + B u_t \\
\end{aligned}
\label{eq:state_space}
\end{equation}
where $x_t = [e_{y}, \dot{e}_{y}, e_{\psi}, \dot{e}_{\psi}]^T$ and $u_t = \delta$.
The optimal state feedback control then is derived as a quadratic programming problem that minimizes a quadratic cost function $J$ as
\begin{equation}
\begin{aligned}
   J = \int_{0}^{\infty} ( x_t^T Q x_t + u_t^T R u_t ) dt \\
\end{aligned}
\label{eq:LQR_cost}
\end{equation}
with gain matrices $Q > 0, R > 0$. The resulting optimal control output $u^*_t = K_t x_t$ is computed with the LQR gain $K_t$. The original LQR calculates the control gain by solving the algebraic Riccati equation \citep{lewis2012optimal}. However, the computational cost of solving the equation online is not negligible. Therefore, we derived $K_t$ by solving the Riccati equation offline via the Explicit LQR algorithm \citep{spisak2022robust}. The solutions are then used for initializing a set of LQR gains. Therefore, our controller queries a feasible $K_t$ with respect to the current velocity, $V_t$, without an expensive online optimization process.

The longitudinal model can be described in terms of the nonlinear traction force and aerodynamic drag acting on the ego vehicle. However, since the ego vehicle generates traction force from its engine-based powertrain, it is challenging to model the longitudinal motion with a single dynamic model. Therefore, we designed the longitudinal model to be hierarchical, with a high-level drive-train model and a low-level engine-based powertrain model following the ideas from \citep{kabzan2020amz} and \citep{rajamani2011vehicle}.

\subsubsection{Real-time Perception based Emergency Controller}
\label{sec:real_time_perception_control}
\begin{figure}[t!]
\centerline{\includegraphics[width=.75\textwidth]{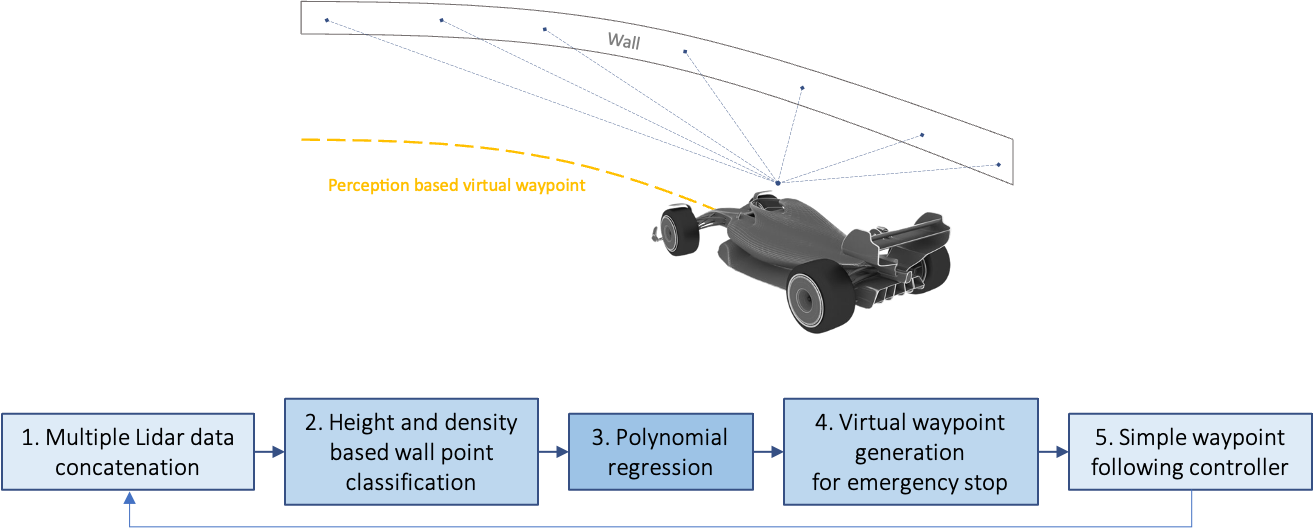}}
\caption[Conceptual visualization and algorithm flow of real-time perception-based lateral vehicle controller.]{Conceptual visualization and algorithm flow of real-time perception-based lateral vehicle controller.} \label{fig:perception_controller}
\end{figure}

Most vehicle controllers use global localization and vehicle status, and the performance of localization directly impacts vehicle control performance. A race vehicle driving at high speed in an uncertain localization condition, such as when the GPS sensor responsible for position recognition fails, or the localization result diverges, requires the ability to safely drive (or stop) the vehicle. To this end, our control stack includes a real-time perception-based controller that functions regardless of the current state of localization. As the name implies, the controller employs only real-time environmental sensor data (in this case, LiDARs) to calculate the path the vehicle will take in body coordinates in real-time and generates a control signal to follow. This controller is engaged only when the system manager detects a fatal failure or bad quality of the localization module. The failure modes of localization include GPS signal disconnection or a high estimation of localization uncertainty. The control's conceptual visualization and algorithmic order are depicted in Fig. \ref{fig:perception_controller}.

Our key idea is to generate the virtual local path parallel to the track boundary and safely stop the vehicle without a crash. We used preprocessed LiDAR point cloud (introduced in Section \ref{sec:detection}) and conducted polynomial fitting. To minimize fitting error, we chose a model with a smaller fitting error among the first and third-order polynomial models. To minimize the lateral maneuvering in emergency stop scenarios, the virtual path begins from (0,0) in body coordinate if the distance from the wall is over 1.5 m. On the other hand, if the distance to the wall is less than 1.5 m, the virtual path will be generated parallel to the wall with a 1.5 m lateral bias. A pure pursuit controller is in charge of calculating the lateral control. 

\subsubsection{Engine Torque Map based Longitudinal Controller}
\begin{figure}[t!]
\centerline{\includegraphics[width=1.\textwidth]{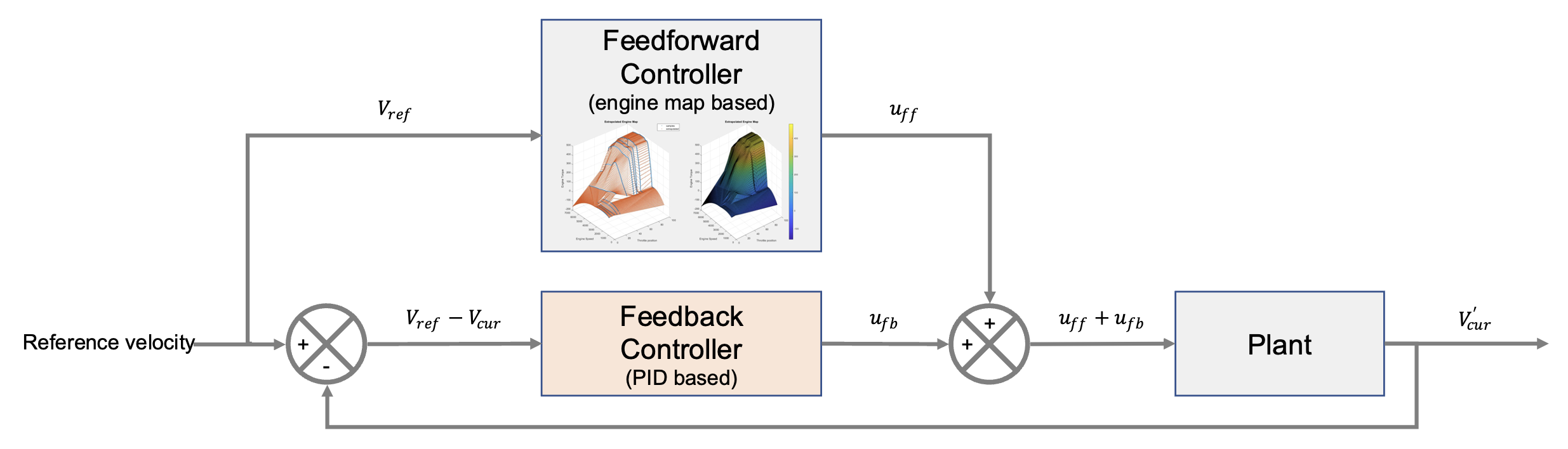}}
\caption[Diagram of a low-level longitudinal controller that outputs the accelerator pedal position from the reference velocity.]{Diagram of a low-level longitudinal controller that outputs the accelerator pedal position from the reference velocity.} \label{fig:lowlevelcontrol}
\end{figure}

Desired acceleration should finally be converted into throttle and brake positions, inputs of the drive-by-wire module. The longitudinal control diagram is shown in Fig. \ref{fig:lowlevelcontrol}.

As shown in the diagram, the feedforward control part computes throttle position based on the engine torque map obtained from the chassis dynamometer testing. In designing the feedforward control module, we make the following assumptions to simplify the problem:
\begin{enumerate}
    \item \textbf{Velocity of the vehicle is decided only by engine and brake forces.} Only the throttle and brakes are required to control the vehicle's speed while driving. The track's geometry (e.g., slope, bank) and environmental conditions (e.g., wind) can be ignored. 
    \item \textbf{The torque converter is fully locked.} Torque from the engine passes directly through the transmission without loss.
    \item \textbf{The tire slip is negligible.} There is no tire slip if the low-level follows the desired velocity.
\end{enumerate}
Deceleration and control errors are corrected using a simple PID-based feedback controller.

\section{Experimental Results}
\label{sec:evaluation}
\subsection{High-speed Solo Lap}
\label{sec:High Speed Field Testing Result:Solo Lap} 
\begin{figure}[b!]
\centerline{\includegraphics[width=1.\textwidth]{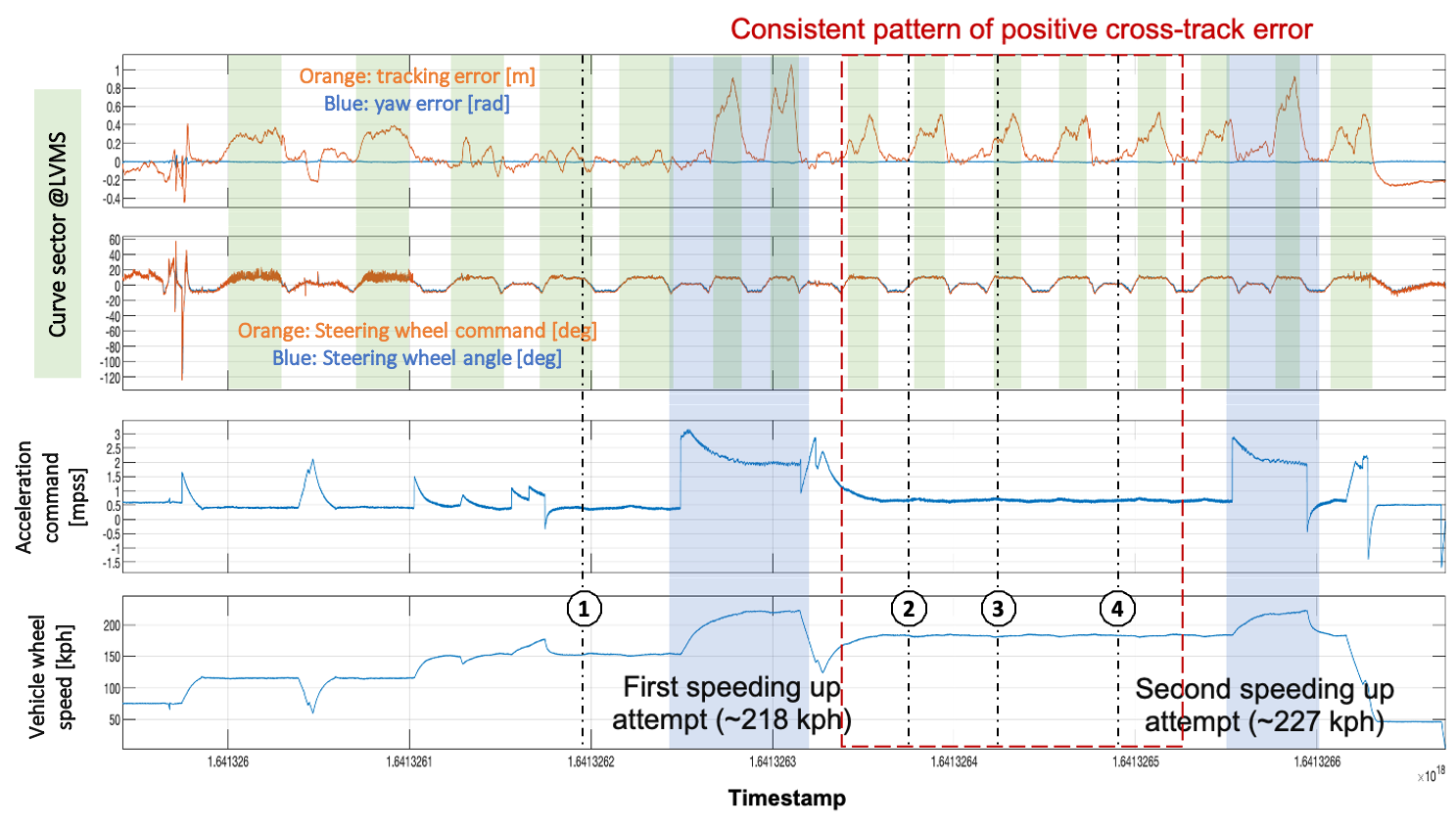}}
\caption[High-speed solo driving experimental results. (First row) Orange and blue lines represent the lateral deviation and yaw error with respect to the reference line. (Second row) Orange and blue lines represent the steering command and position in degree, respectively.]{High-speed solo driving experimental results. (First row) Orange and blue lines represent the lateral deviation and yaw error with respect to the reference line. (Second row) Orange and blue lines represent the steer command and position in degree, respectively.} \label{fig:performance-control-plot}
\end{figure}

\begin{figure}[t!]
\centerline{\includegraphics[width=.83\textwidth]{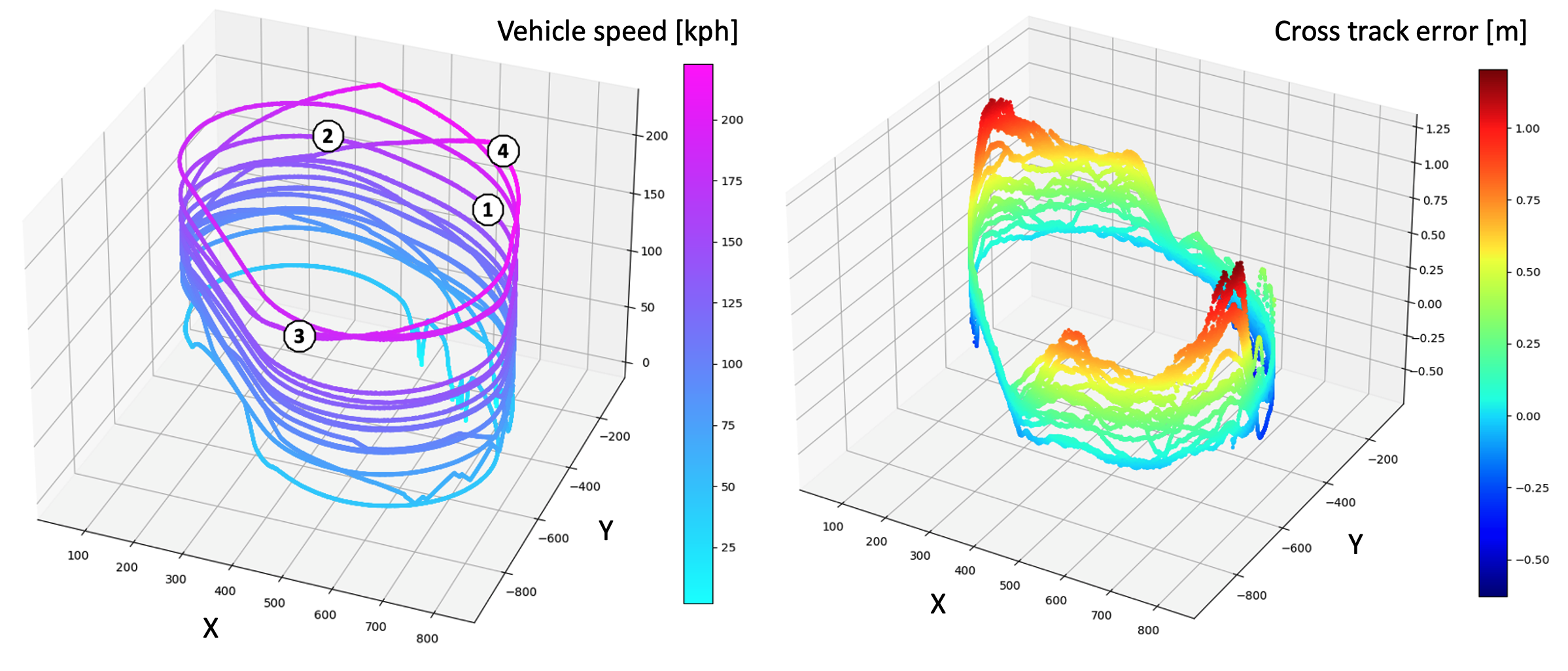}}
\caption[Lateral deviation from the race line plot according to the driven trajectory. The numbers overlaid in the left figure match with timestamps of Fig. \ref{fig:performance-control-plot}.]{Lateral deviation from the race line plot according to the driven trajectory. The numbers overlaid in the left figure match with timestamps of Fig. \ref{fig:performance-control-plot}.} \label{fig:performance-3d-cte}
\end{figure}

This section presents the field testing results of high-speed driving conducted on January 4, 2022, to validate the system's stability before the race event in LVMS. The maximum speed was set to 235 ${km}/{h}$, while the maximum acceleration was set to 3 ${m}/{s^2}$. The experiment was conducted while gradually increasing the maximum speed. 
The control result at the time is depicted in Fig. \ref{fig:performance-control-plot}.
In addition, Fig. \ref{fig:performance-3d-cte} shows the driving speed and lateral deviation.


\begin{figure}[t!]
\centerline{\includegraphics[width=.83\textwidth]{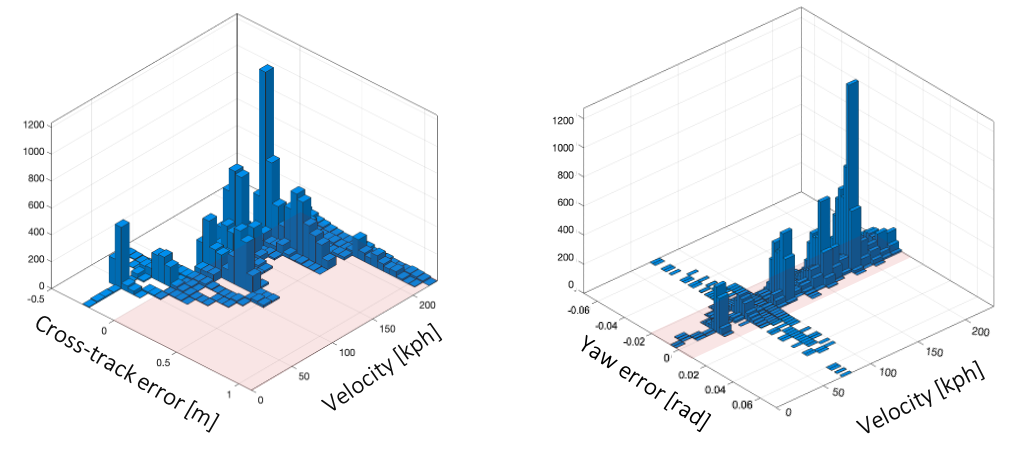}}
\caption[Tracking error results. (Left) Lateral deviation (Right) Yaw error.]{Tracking error results. (Left) Lateral deviation (Right) Yaw error.} \label{fig:performance-histo}
\end{figure}

Fig. \ref{fig:performance-histo} shows the lateral deviation from the race line. The maximum cross-track error was about 1.15m at speeds over 220 ${km}/{h}$. Furthermore, 87\% of the cross-track errors showed positive signs, which means that our vehicle was on the left side of the race line. This is because the DALLARA-AV21 has an oval setting (left camber setting) which is not modeled in our vehicle model. Also, the yaw error is depicted in Fig. \ref{fig:performance-histo} on the left. There was a considerable yaw error at 75 ${km}/{h}$ because of switching from the pit-out path to the race line on the track. In other cases, 97\% of yaw errors are kept within the 0.02 rad. The quantitative analysis results of a high-speed lap are summarized in Table \ref{tab:performance-lab-analysis}. In this experiment, we could drive at a maximum speed of 227.2 ${km}/{h}$.

\begin{table}[h!]
\caption[High-speed driving performance analysis result.]{High-speed driving performance analysis result.}
\label{tab:performance-lab-analysis}
\centering
\begin{tabular}{cccccc}
\hline
\multirow{3}{*}{{\begin{tabular}[c]{@{}c@{}}Velocity\\ range  {[}${km}/{h}${]}\end{tabular}}} & \multicolumn{3}{c}{{Tracking error {[}m{]}}}& \multirow{3}{*}{{\begin{tabular}[c]{@{}c@{}}Max yaw\\ error {[}rad{]}\end{tabular}}} & \multirow{3}{*}{{\begin{tabular}[c]{@{}c@{}}Max driving\\ speed {[}${km}/{h}${]}\end{tabular}}} \\ \cline{2-4}& {Max} & {\begin{tabular}[c]{@{}c@{}}Absolute\\ mean\end{tabular}} & {\begin{tabular}[c]{@{}c@{}}Standard\\ deviation\end{tabular}} &&\\ \hline \hline
v \textless 100& 0.59& 0.15& 0.192& 0.074& \multirow{3}{*}{227.2}\\ \cline{1-5}
100 \textless v \textless 150& 0.71& 0.17& 0.214& 0.07&\\ \cline{1-5}
150 \textless v& 1.15& 0.3& 0.443& 0.067&\\ \hline
\end{tabular}
\end{table}

\subsection{1:1 Head-to-head Race Event}
\label{sec:1:1 Head-to-head Race} 
This section introduces the race results at the IAC's quarter-final and semi-final rounds. A total of 5 teams that passed qualification participated in the competition: \emph{TUM} (Germany), \emph{KAIST} (South Korea), \emph{PoliMOVE} (Italy), \emph{TII EuroRacing} (Italy), and \emph{Auburn} (USA). The race was held in a tournament format. The team who successfully overtook the opponent advanced to the next round. \emph{PoliMOVE} was given a bye because they demonstrated the fastest driving during qualifications. \emph{Auburn} was our quarter-final rival. Race control set the starting defender's target velocity at 128 ${km}/{h}$ (80 mph). We were instructed by race control to begin the race as an attacker, while \emph{Auburn} started as a defender. To ensure safety, the two vehicles exit the pit box in the order of defender and attacker. 

\begin{figure}[b!]
\centerline{\includegraphics[width=1.0\textwidth]{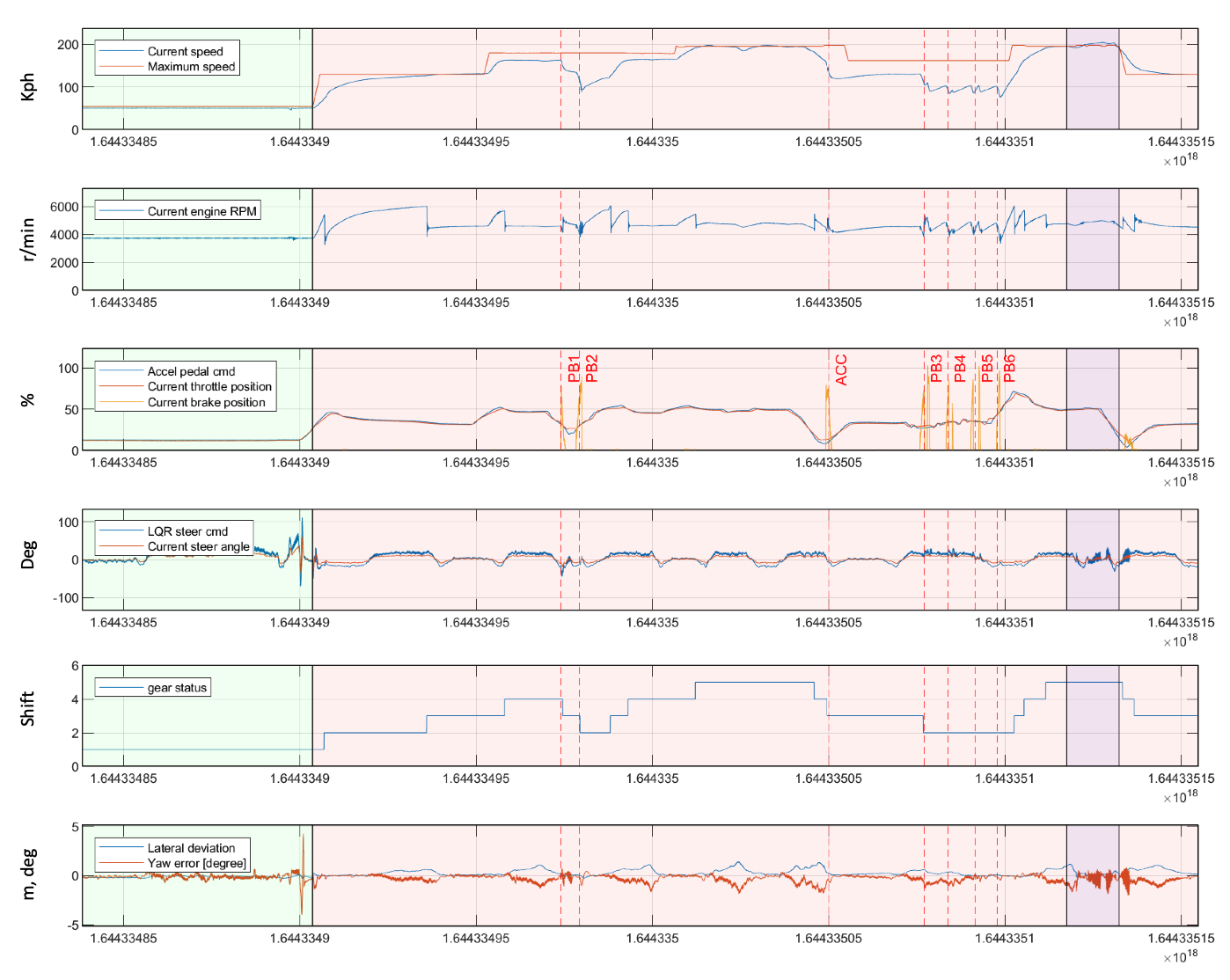}}
\caption[Quarter-final driving result. Green areas indicate pit-outs. The formation lap was performed according to communication with race control, indicated by the pink area. A purple area indicates the section overtaking a vehicle traveling at 128 ${km}/{h}$ (80 mph), and the driving speed at that time was about 212 ${km}/{h}$.]{Quarter-final driving result. Green areas indicate pit-outs. The formation lap was performed according to communication with race control, indicated by the pink area. A purple area indicates the section overtaking a vehicle traveling at 128 ${km}/{h}$ (80 mph), and the driving speed at that time was about 212 ${km}/{h}$.} \label{fig:auburn_plot_full}
\end{figure}

The control result from the pit out in the quarter-final round is shown in Fig. \ref{fig:auburn_plot_full}. We began the race as an attacker attempting to overtake the opposing team. We performed pit-out and formation laps based on race control signals. A green area denotes the section where pit-out was done. During the pit-out mission, the speed was around 50 ${km}/{h}$. As soon as the pitting out is completed, our vehicle aligns with the race line. The formation lap following the race control was shown in the pink area.

\begin{figure}[t!]
\centerline{\includegraphics[width=1.0\textwidth]{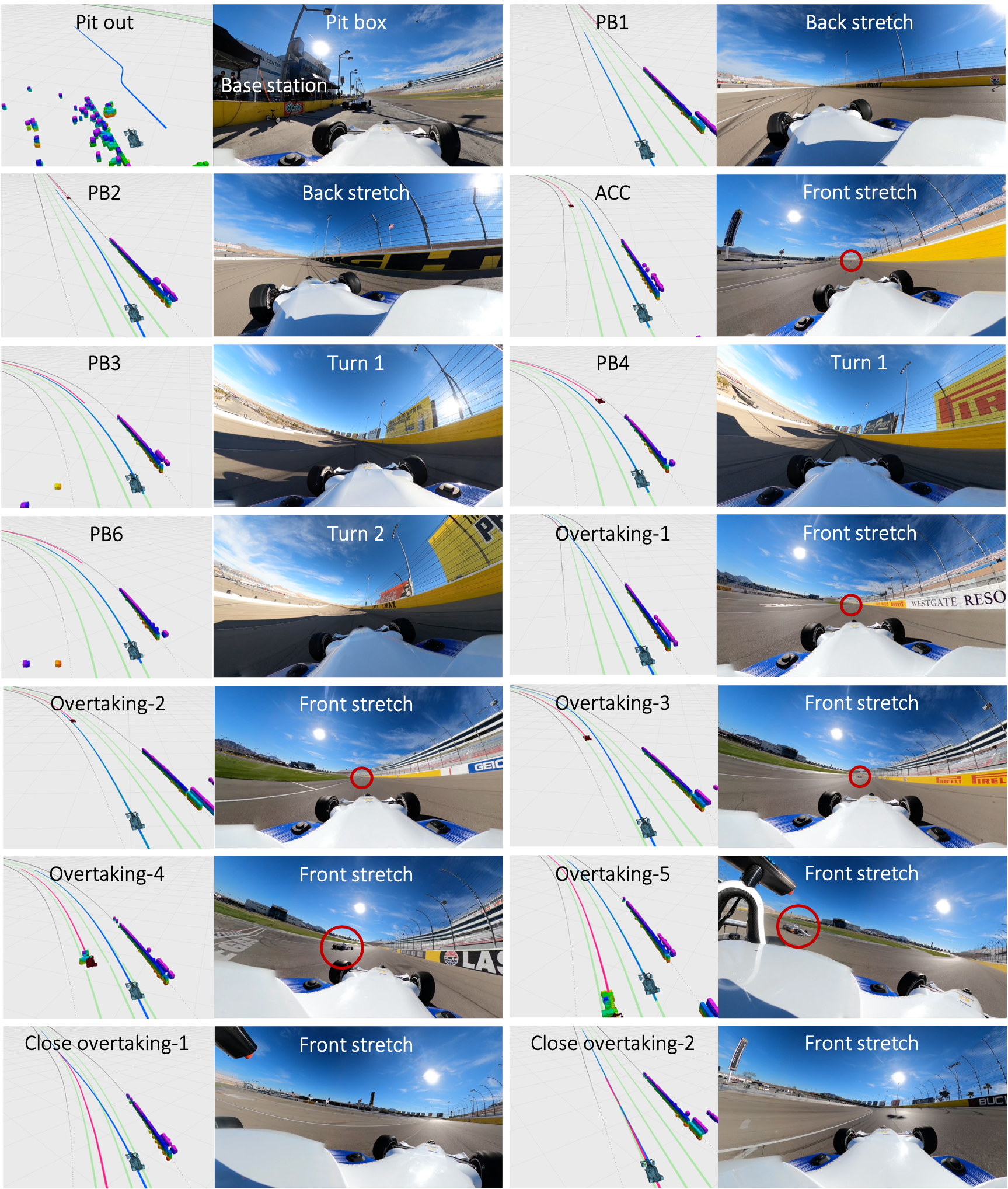}}
\caption[Onboard camera view with perception and planning result visualization results at quarter-final race.]{Onboard camera view with perception and planning result visualization results at quarter-final race.} \label{fig:auburn_cam}
\end{figure}
\cleardoublepage

Following the race control, the attacker should reduce the gap between himself and the defender. We communicated with the race control during the formation lap and increased the maximum speed. In Fig. \ref{fig:auburn_plot_full}, PB1 indicates when the false detection (hereafter referred as to FD) firstly occurred, and unnecessary deceleration was performed. At the time, our vehicle was running the formation lap following the inner side of the track, and a FD from the perception module occurred near the right track boundary. At the time of PB2, the race control provided an attacker flag, allowing us to attempt overtaking, and our planner switched to the outer line, the attacker mode's default line. Simultaneously, second phantom braking occurred due to a FD. We increased the attacker mode's speed to 198 ${km}/{h}$ to quickly catch the opponent. At ACC, \emph{Auburn}'s vehicle was accurately detected for the first time at a distance of about 100m and decelerated to the defender speed of 128 ${km}/{h}$ (80 mph) to maintain the distance. After a while, four times phantom brakes (PB3, PB4, PB5, and PB6) occurred, and the speed dropped to about 100 ${km}/{h}$ (60 mph). After the FD disappeared, our vehicle accelerated again up to our top speed, 200 ${km}/{h}$ (124 mph), following the race line. When the distance gap between the two vehicles was about 34 m in the front stretch, our speed was approximately 185 ${km}/{h}$ (115 mph), and we started overtaking by safely switching the path to the outer line. After passing, we returned to the inner track line about 32 m of safety margin from \emph{Auburn}'s vehicle. The above-mentioned situations are shown in Fig. \ref{fig:auburn_cam} in order of time. We performed defender-attacker role switching after our overtaking, but the race control halted the race due to an error in the \emph{Auburn}'s system.

\begin{figure}[t!]
\centerline{\includegraphics[width=.9\textwidth]{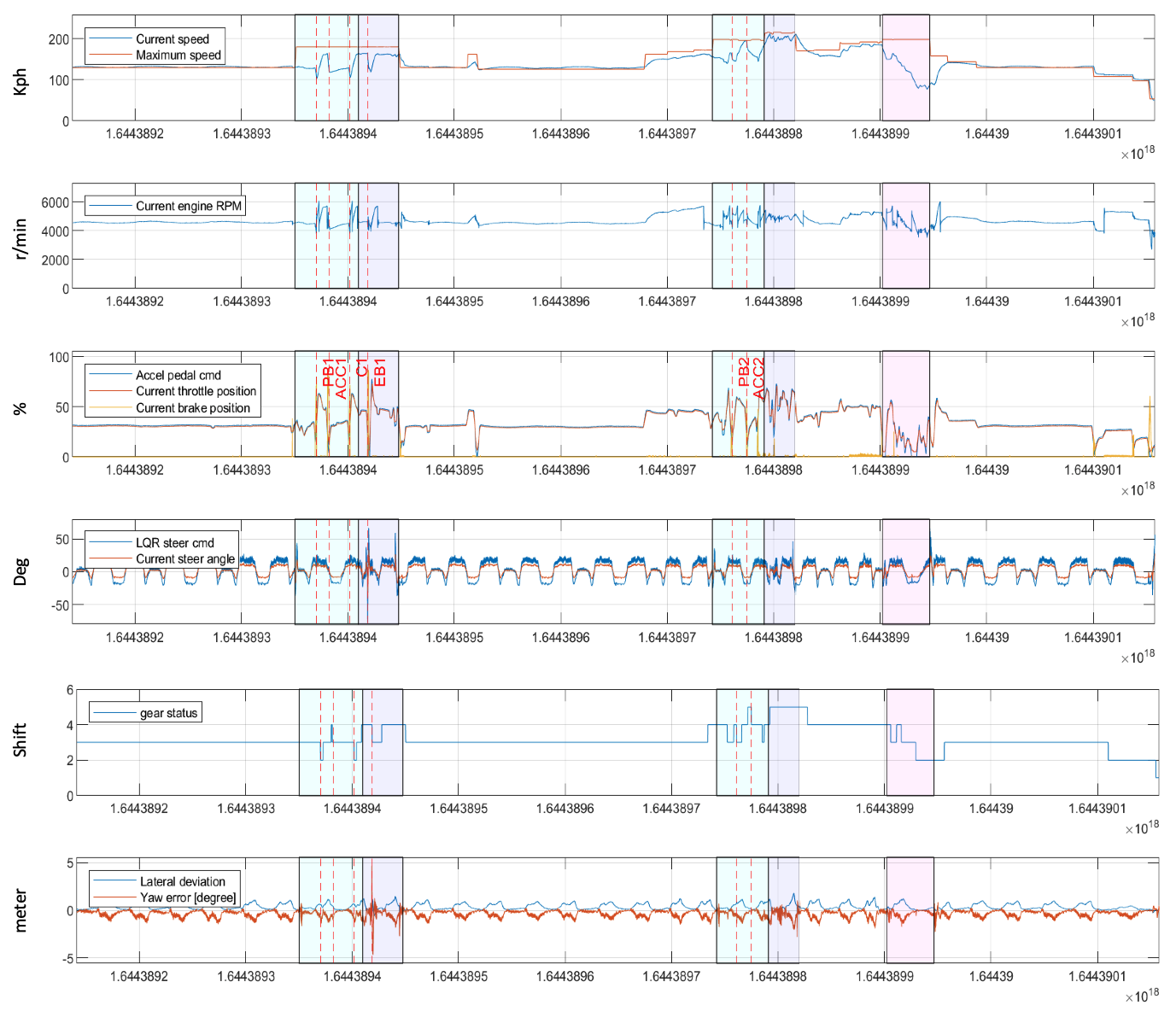}}
\caption[Semi-final driving result. Overtaking in attacker mode (80 mph, 100 mph) was performed a total of 2 times, and defender role (80 mph, 100 mph, 115 mph) was performed three times. All missions of pit-in and out, formation lap, and race were performed, and driving was performed for about 18 minutes. When overtaking a vehicle traveling at 100 mph, the top speed was about 204 ${km}/{h}$. FDs caused repeated phantom braking. We decided that it was not safe for both teams, so we finally asked the race control a black flag for a pit-in, and the pink area shows the point in time.]{Semi-final driving result. Overtaking in attacker mode (80 mph, 100 mph) was performed a total of 2 times, and defender role (80 mph, 100 mph, 115 mph) was performed three times. All missions of pit-in and out, formation lap, and race were performed, and driving was performed for about 18 minutes. When overtaking a vehicle traveling at 100 mph, the top speed was about 204 ${km}/{h}$. FDs caused repeated phantom braking. We decided it was unsafe for both teams, so we finally asked the race control a black flag for a pit-in, and the pink area shows the point in time.} \label{fig:polimove_plot_full}
\end{figure}
\begin{figure}[t!]
\centerline{\includegraphics[width=.76\textwidth]{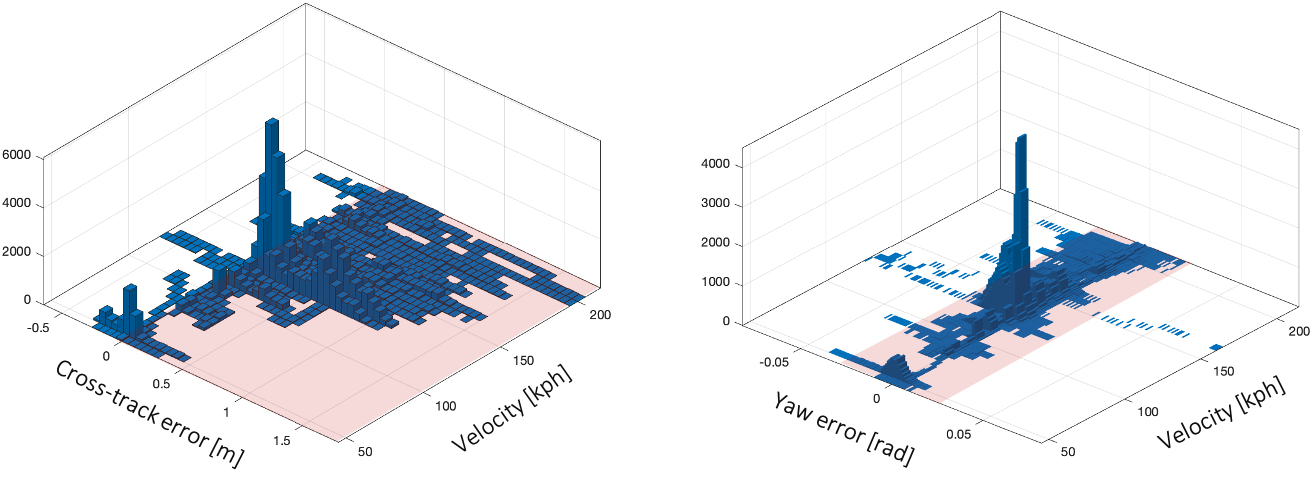}}
\caption[Tracking error histogram according to the driving speed during the CES2022 head-to-head autonomous race. (Left) Lateral deviation (Right) Yaw error.]{Tracking error histogram according to the driving speed during the CES2022 head-to-head autonomous race. (Left) Lateral deviation (Right) Yaw error.} \label{fig:h2h-histo}
\end{figure}
\begin{figure}[t!]
\centerline{\includegraphics[width=.8\textwidth]{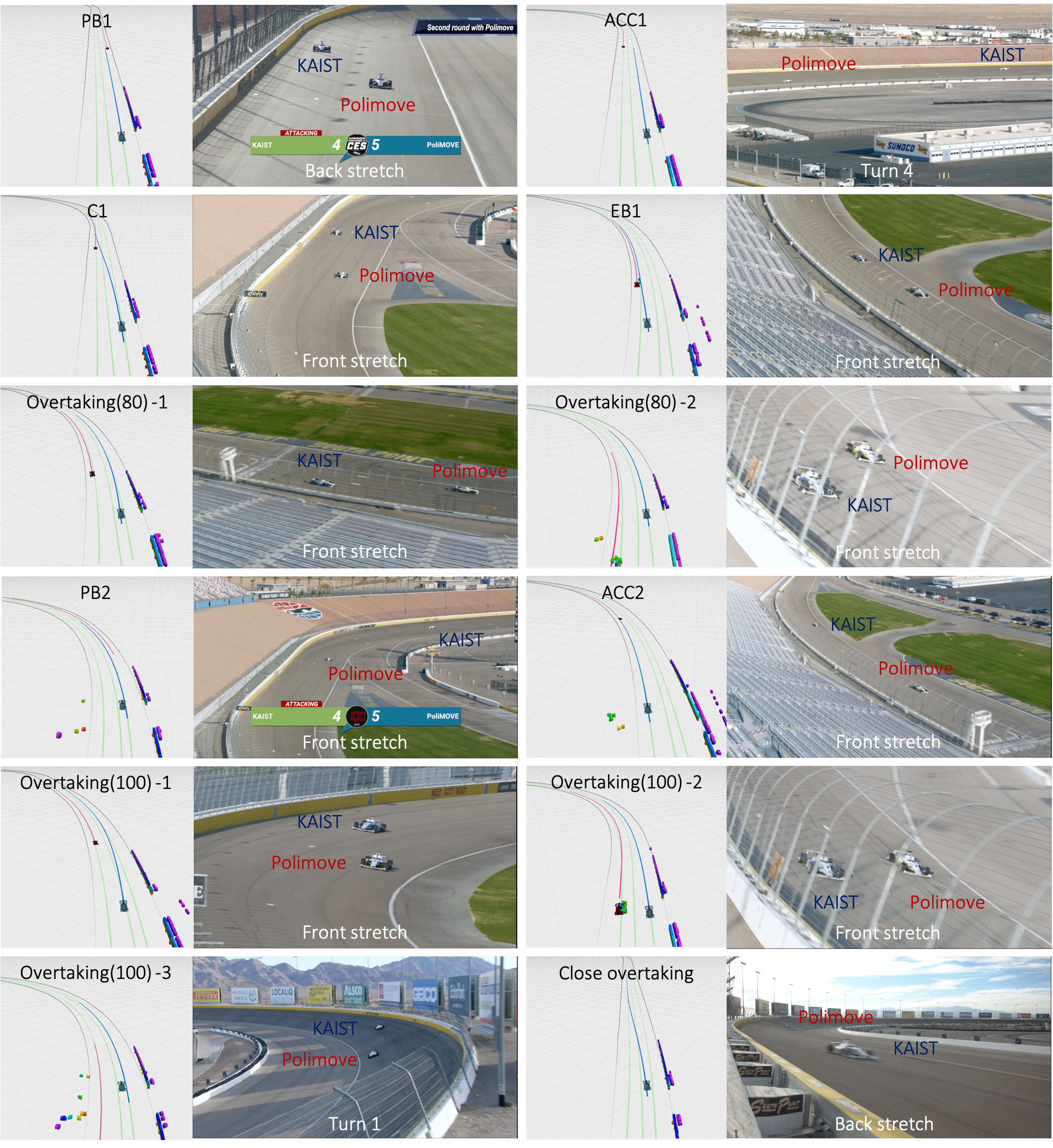}}
\caption[Onboard camera view with perception and planning result visualization results at semi-final race.]{ Onboard camera view with perception and planning result visualization results at semi-final race.} 
\label{fig:semi-final-view}
\end{figure}

As in the quarter-final, we performed a total of three times of defender and two times of attacker in the semi-final with \emph{PoliMOVE}. Fig. \ref{fig:polimove_plot_full} shows the overall control results in the semi-final. We also experienced phantom braking several times, similar to the quarter-finals. The first time we faced the phantom braking is shown by PB1 in the third row of Fig. \ref{fig:polimove_plot_full}. After the FD disappeared, our vehicle slowed down (ACC1) to maintain the distance while waiting for the front stretch overtaking zone. C1 timestamp shows when the vehicle got the overtaking flag from the race control and switched to the race line since there was no collision within 2 seconds. At EB1, our vehicle started to overtake the front vehicle at a speed of 150 ${km}/{h}$ (See Fig. \ref{fig:semi-final-view}, Overtaking(80)-1, and Overtaking(80)-2). The next round of defender speed was set to 160 ${km}/{h}$ (100 mph). PB2 is a phantom braking point that occurred while closing the distance with the \emph{PoliMOVE}, and we accelerated up to 204 ${km}/{h}$ (126 mph), successfully overtaking an opposing vehicle traveling at 160 ${km}/{h}$ (100 mph). We started overtaking at the end of the front stretch due to phantom braking, and both vehicles were side-by-side at turn 1. Our planner did not permit closing overtaking in the high banking angle zone and completed the overtaking maneuver at the backstretch. However, we decided to request a black flag to race control for the retirement because of repeated phantom braking (shown in the pink area of Fig. \ref{fig:polimove_plot_full}). 

In the same way as the method introduced in Section \ref{sec:High Speed Field Testing Result:Solo Lap}, the control performance in the head-to-head race was analyzed. Fig. \ref{fig:h2h-histo} shows the lateral displacement and yaw error. Fig. \ref{fig:h2h-histo} appears similar to Fig. \ref{fig:performance-histo}, which means that our controller can accurately track the fixed path but also dynamically changing path. The quantitative analysis results are summarized in Table \ref{tab:h2h-race-analysis}. Our maximum speed during the semi-final was 212 ${km}/{h}$ (132 mph), and accelerations up to 12.41 ${m}/{s^2}$. 
The root cause of all FDs was a mismatch of the right side of the track boundary, especially at the curve sectors. Thus, our geofence filter with a 1 m threshold distance sometimes could not correctly remove the LiDAR-based clustering noise or raw radar data.

\begin{table}[b!]
\centering
\caption[Head-to-head race analysis result. (ABYE: Absolute max yaw error. MLA: Max longitudinal acceleration. AMLA: Absolute max lateral acceleration. MS: Max speed.)]{Head-to-head race analysis result. (ABYE: Absolute max yaw error. MLA: Max longitudinal acceleration. AMLA: Absolute max lateral acceleration. MS: Max speed.)}
\label{tab:h2h-race-analysis}
\begin{tabular}{cccccccc}
\hline
 &
  \multicolumn{3}{c}{Tracking error ↓ {[}m{]}} &
  {\color[HTML]{0D0D0D} } &
  {\color[HTML]{0D0D0D} } &
  {\color[HTML]{0D0D0D} } &
  {\color[HTML]{0D0D0D} } \\ \cline{2-4}
\multirow{-2}{*}{\begin{tabular}[c]{@{}c@{}}Velocity range\\ {[}${km}/{h}${]}\end{tabular}} &
  {\color[HTML]{0D0D0D} Max} &
  {\color[HTML]{0D0D0D} \begin{tabular}[c]{@{}c@{}}Absolute\\ mean\end{tabular}} &
  {\color[HTML]{0D0D0D} \begin{tabular}[c]{@{}c@{}}Standard\\ deviation\end{tabular}} &
  \multirow{-2}{*}{{\color[HTML]{0D0D0D} \begin{tabular}[c]{@{}c@{}}ABYE\\ {[}rad{]}\end{tabular}}} &
  \multirow{-2}{*}{{\color[HTML]{0D0D0D} \begin{tabular}[c]{@{}c@{}}MLA\\ {[}${m}/{s^2}${]}\end{tabular}}} &
  \multirow{-2}{*}{{\color[HTML]{0D0D0D} \begin{tabular}[c]{@{}c@{}}AMLA\\ {[}${m}/{s^2}${]}\end{tabular}}} &
  \multirow{-2}{*}{{\color[HTML]{0D0D0D} \begin{tabular}[c]{@{}c@{}}MS\\ {[}${km}/{h}${]}\end{tabular}}} \\ \hline \hline
{\color[HTML]{0D0D0D} v \textless 100} &
  {\color[HTML]{0D0D0D} 0.4005} &
  {\color[HTML]{0D0D0D} 0.1068} &
  {\color[HTML]{0D0D0D} 0.0995} &
  {\color[HTML]{0D0D0D} 0.0415} &
  {\color[HTML]{0D0D0D} } &
  {\color[HTML]{0D0D0D} } &
  {\color[HTML]{0D0D0D} } \\ \cline{1-5}
{\color[HTML]{0D0D0D} 100 \textless v \textless 150} &
  {\color[HTML]{0D0D0D} 1.2318} &
  {\color[HTML]{0D0D0D} 0.3846} &
  {\color[HTML]{0D0D0D} 0.2905} &
  {\color[HTML]{0D0D0D} 0.0815} &
  {\color[HTML]{0D0D0D} } &
  {\color[HTML]{0D0D0D} } &
  {\color[HTML]{0D0D0D} } \\ \cline{1-5}
{\color[HTML]{0D0D0D} 150 \textless v} &
  {\color[HTML]{0D0D0D} 1.7791} &
  {\color[HTML]{0D0D0D} 0.4872} &
  {\color[HTML]{0D0D0D} 0.3606} &
  {\color[HTML]{0D0D0D} 0.0812} &
  \multirow{-3}{*}{{\color[HTML]{0D0D0D} \begin{tabular}[c]{@{}c@{}}12.4155\\ (1.266 g)\end{tabular}}} &
  \multirow{-3}{*}{{\color[HTML]{0D0D0D} \begin{tabular}[c]{@{}c@{}}6.8750\\ (0.701 g)\end{tabular}}} &
  \multirow{-3}{*}{{\color[HTML]{0D0D0D} \begin{tabular}[c]{@{}c@{}}212.5\\ (132 mph)\end{tabular}}} \\ \hline
\end{tabular}x
\end{table}

\subsection{Computational Performance Analysis}
\label{sec:computational-performance}

Real-time systems, relying on either soft or hard timing constraints, require proper diagnostics to verify that the design constraints are respected at run-time. Timing analysis for complex systems is an active research topic. Recent techniques, including some specifically designed for ROS2 \citep{li2022autowareperf,bedard2022ros2tracing}, have been released only recently. In this section, an analysis of the timing properties of our proposed autonomy stack is given. The timing measurements have been carried out on the platform detailed in Table \ref{tab:timing:platform}, by replaying recorded data. The tracing tool adopted to measure the software performance is derived from \texttt{Autoware\_perf} \citep{li2022autowareperf}, which in turn is based on the \texttt{ros2\_tracing} \citep{bedard2022ros2tracing} and LTTng \citep{desnoyers2006lttng} tools. 

\begin{figure}[t!] 
    \begin{subfigure}{0.333\textwidth}
        \includegraphics[width=\textwidth]{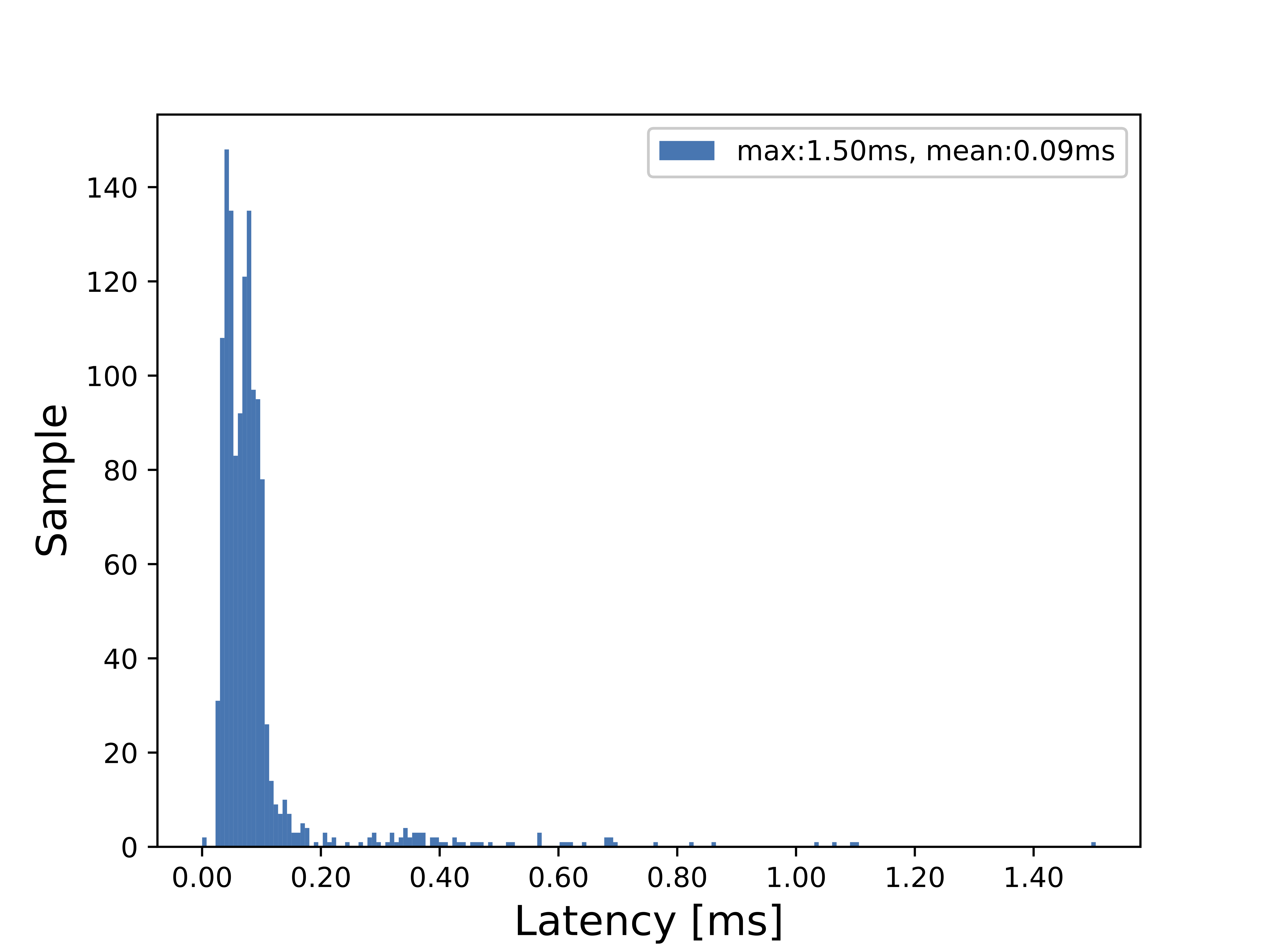}
        \caption{State estimation module.} \label{fig:timing:awl}
    \end{subfigure}
    \begin{subfigure}{0.333\textwidth}
        \includegraphics[width=\textwidth]{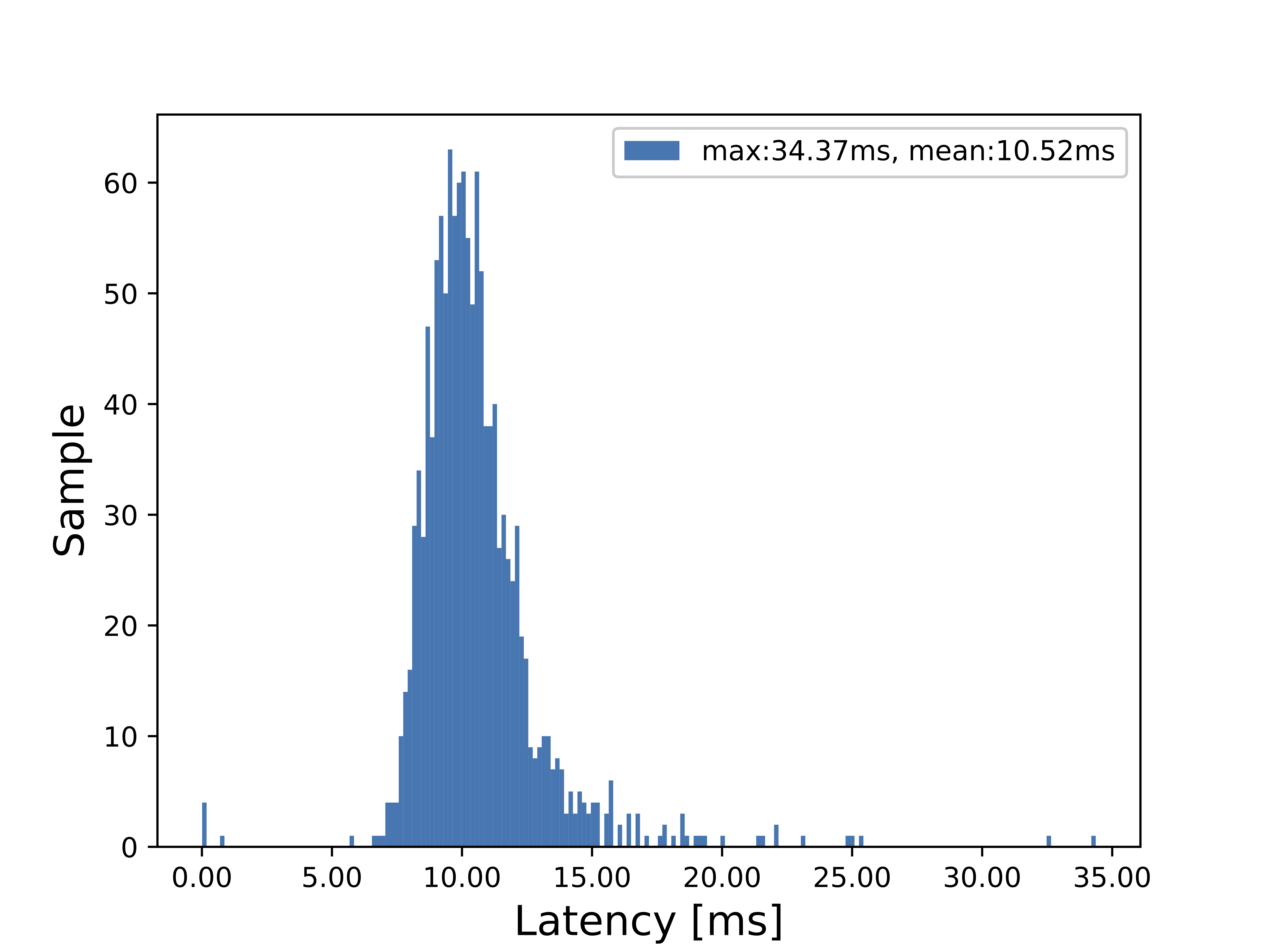}
        \caption{LiDAR clustering module.} \label{fig:timing:clu}
    \end{subfigure}
    \begin{subfigure}{0.333\textwidth}
        \includegraphics[width=\textwidth]{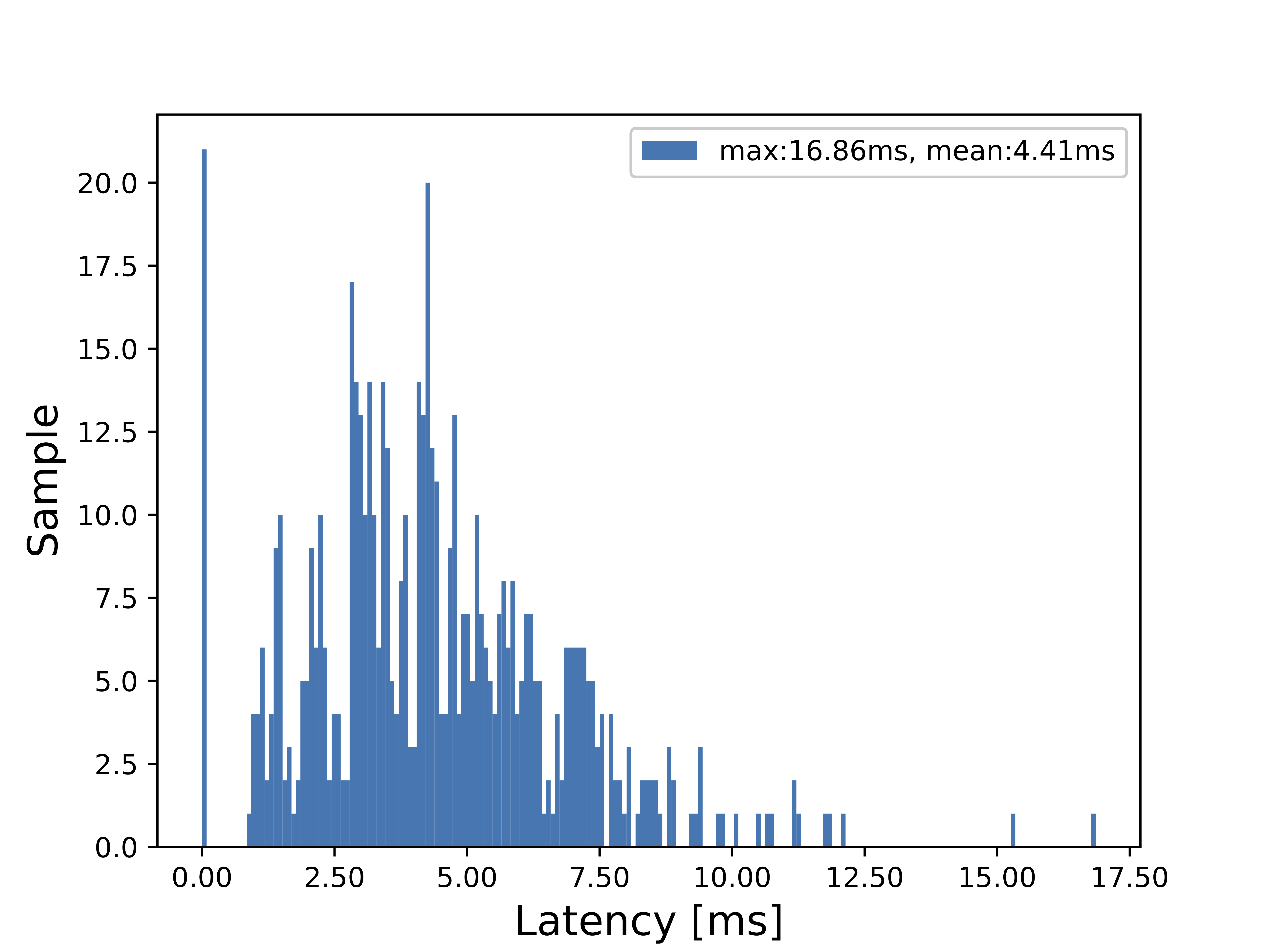}
        \caption{Geofence filtering module.} \label{fig:timing:gff}
    \end{subfigure}
    
    \begin{subfigure}{0.333\textwidth}
        \includegraphics[width=\textwidth]{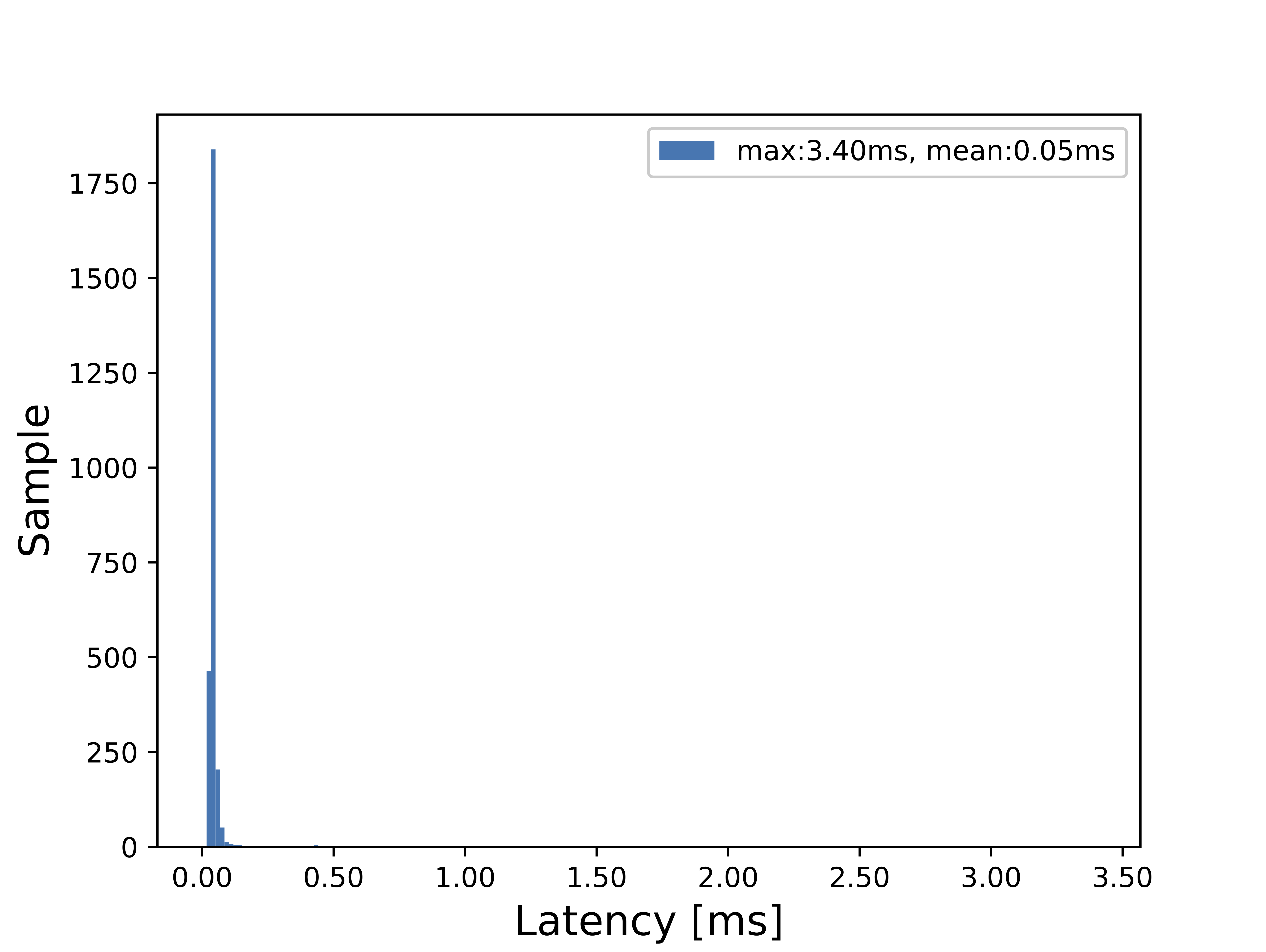}
        \caption{Overtaking planner module.} \label{fig:timing:dyn}
    \end{subfigure}
    \begin{subfigure}{0.333\textwidth}
        \includegraphics[width=\textwidth]{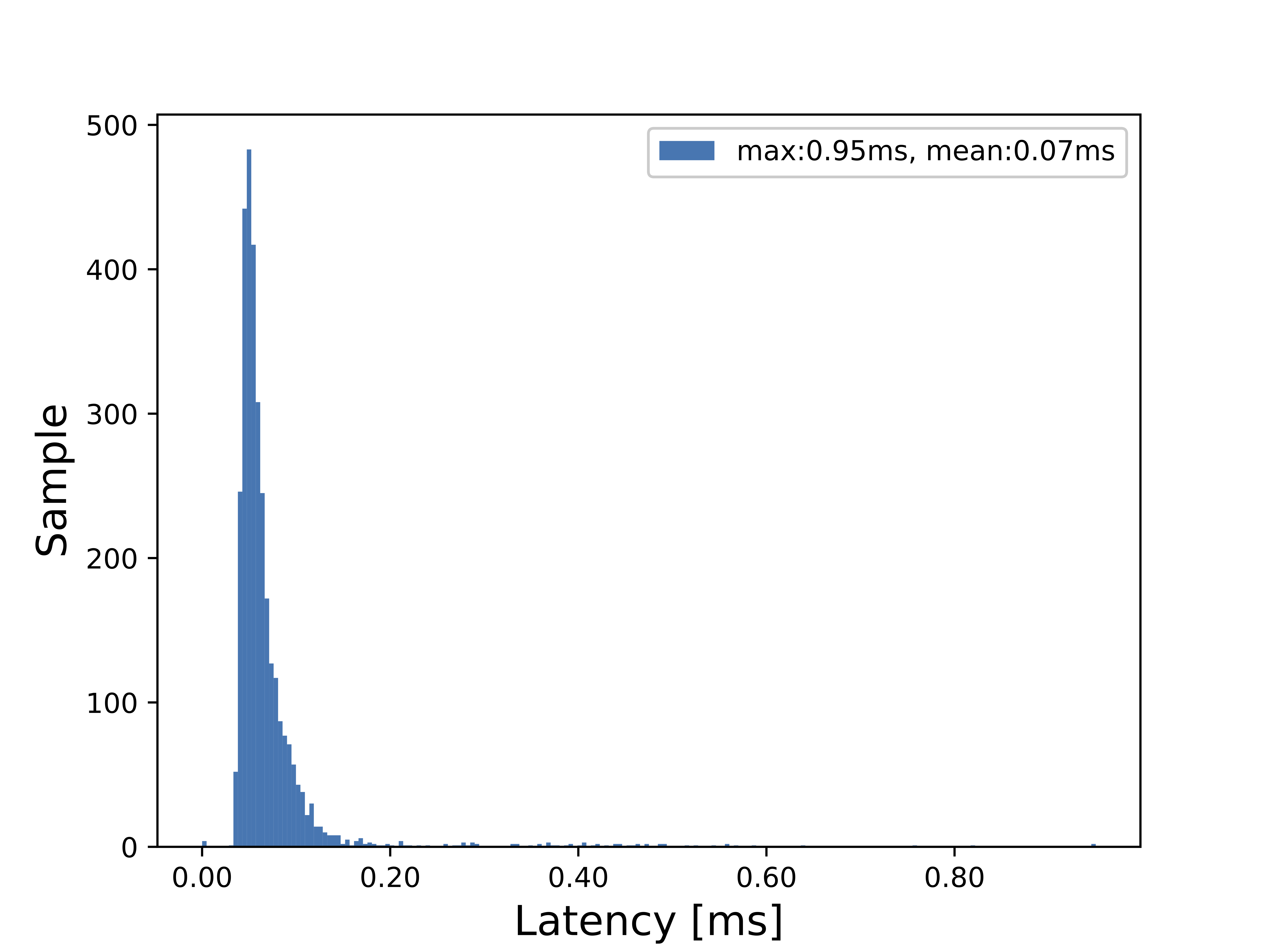}
        \caption{LQR controller module.} \label{fig:timing:lqr}
    \end{subfigure}
    \begin{subfigure}{0.333\textwidth}
        \includegraphics[width=\textwidth]{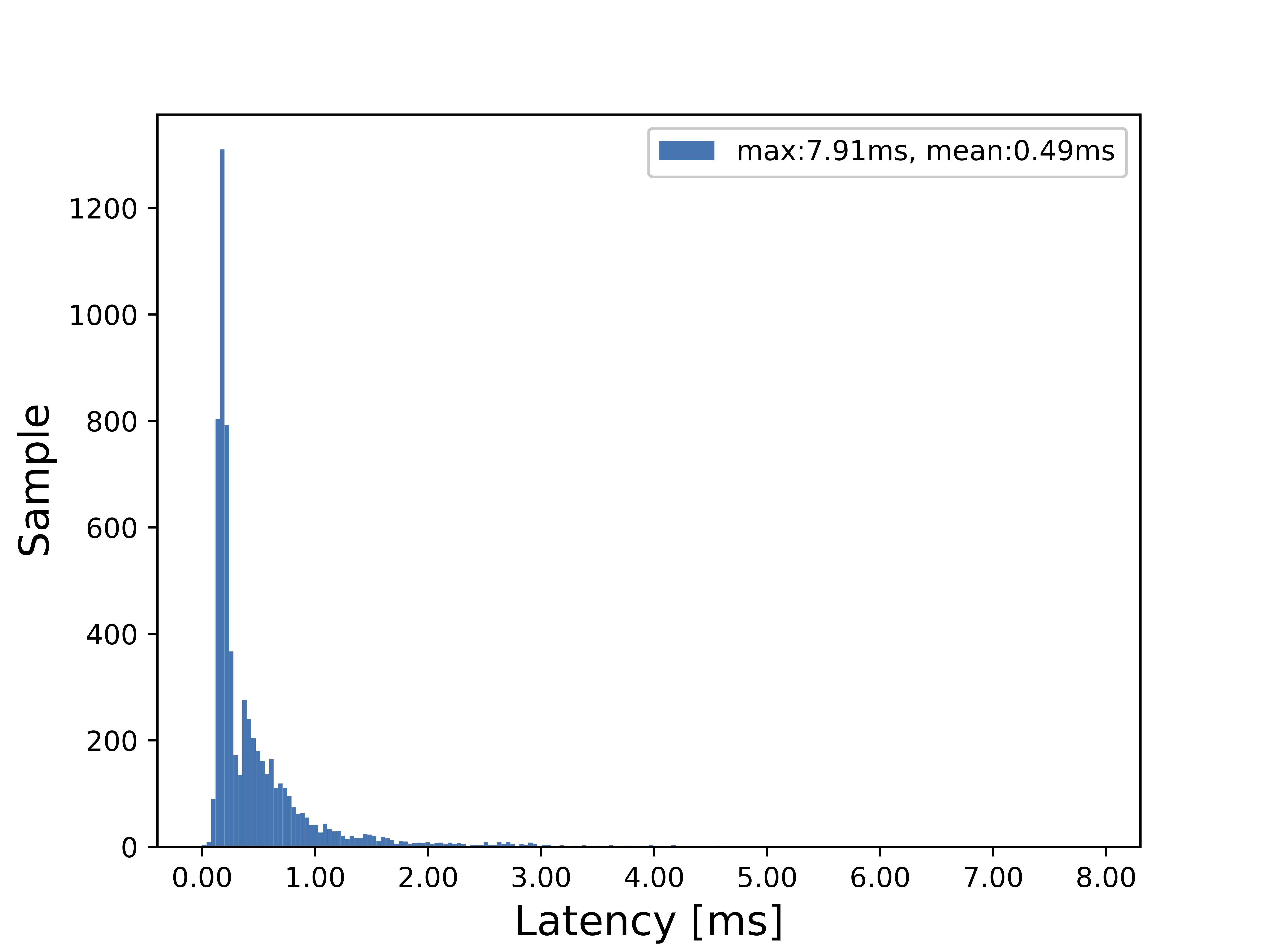}
        \caption{System status manager module.} \label{fig:timing:ssm}
    \end{subfigure}
    \caption{Latency analysis results.} \label{fig:timing}
\end{figure}

We measured the execution time of the functions that compose the main execution pipeline (registered as ROS2 topic or timer callbacks). For the performance analysis, we plot the number of samples over their execution time (discretizing the timeline). Both the evaluation workstation and the computation system on Dallara-AV21 run on the default Linux 5.8 kernel, which employs a Completely Fair Scheduling algorithm \citep{pabla2009cfs}. Fig. \ref{fig:timing} shows various task duration plots.

\begin{table}[b!]
\centering
\caption[Timing analysis workstation's specifications.]{Timing analysis workstation's specifications.}
\label{tab:timing:platform}
\begin{tabular}{cc}
\hline
Component & Specification                            \\ \hline \hline
CPU       & Intel Core i9-20980HK 2.40 GHz (16T, 8C) \\ \hline
GPU       & Nvidia GeForce RTX 3080 Laptop x 1       \\ \hline
RAM       & 32 GB                                    \\ \hline
Storage   & 2TB Samsung 970Evo NVMe SSD              \\ \hline
Software  & ROS2 Galactic on Ubuntu 20.04            \\ \hline
\end{tabular}
\end{table}

The timing requirements set during the design phase (i.e., 100 Hz operation frequency), are met by the stack, according to the data. The nodes composing the localization-planning-control sequence performs well-below the 10 ms deadline, with reasonably low variance (Figures \ref{fig:timing:awl}, \ref{fig:timing:dyn}, \ref{fig:timing:lqr}). Similar conclusions can be drawn with regard to the main horizontal component, the System Status Manager (Fig. \ref{fig:timing:ssm}). The perception stack (intended as the detection-tracking-prediction sequence) shows lower performance compatibly with the higher resource-consuming nature of the task. In particular, the lidar clustering algorithm sets its mean execution time at 10.52 ms, peaking at 34.37 ms (Fig. \ref{fig:timing:clu}), while the geofence filtering function averages at 4.41 ms with a registered maximum of 16.86 ms (Fig. \ref{fig:timing:gff}). Moreover, while the former's variance is reasonably limited, the spread-out shape of the latter's plot suggests that the function's flow might have an irregular duration. This result, although not optimal, sets the worst-case operation frequency of the perception task around 30 Hz, compatibly with the best-case performance of the available laser-based sensors. 

Although the results' scale can be heavily influenced by environmental factors (e.g., hardware and OS configuration, number of concurrent processes, external conditions), the relative proportion of time consumption is well represented by the given data, regardless of the testing environment. As per an evaluation of the end-to-end latency, the data suggest that the average latency sets well-below 10 ms for what concerns the control path (state estimation, planning, control), including the impact of inter-process communication, which averaged around 0.01 ms on our timing analysis workstation.

\section{Discussion and Conclusion}
\label{sec:conclusion}
In this paper, we presented the full-stack autonomous racing software developed by team \emph{KAIST} for the Indy Autonomous Challenge (IAC). Our autonomy solution comprises multi-modal perception, a high-speed overtaking planner, a resilient control stack, and a system status manager. All the subsystems of the proposed autonomy stack are developed following our key design principles, aiming to achieve dependability, evolvability, and performance. Even though our autonomy solution is developed targeting the autonomous racing domain, we believe that our system architecture and design principles can be applied to a wide range of robotic applications, especially when it comes to the high-performance, safety-critical, and high-cost-of-failure application domains. 

The proposed system was integrated into a full-scaled autonomous race car (Dallara AV-21) and extensively validated through field tests and race events. We, team \emph{KAIST}, accomplished every mission (including autonomous pit-in/out, static obstacles avoidance, and obeying race flags) in the IAC at Indianapolis Motor Speedway (IMS) in Oct 2021. During IAC at Las Vegas Motor Speedway (LVMS) in Jan 2022, our autonomy demonstrated high-speed head-to-head racing by reaching speeds over 220 ${km}/{h}$ and accelerations of up to 12.41 ${m}/{s^2}$. Our team was one of three teams that successfully finished both race events without system failures or crashes.


Even though we believe that the proposed autonomy solution and the provided results can provide valuable insights to the field robotics community, there are still some technical gaps when it comes to human-like head-to-head racing scenarios. Our autonomy solution was designed based on the IAC race rules, which allowed us to make some fundamental assumptions. For example, our prediction module simplifies the problem by assuming that the other vehicle will maintain a constant speed and a constant displacement from the inner track boundary. However, this assumption is not valid for human-like races. Also, our trajectory planner follows a sampling-based approach which has an advantage in terms of computation burden. However, our planning algorithm has limited capabilities when handling multiple opponents in a competitive scenario. For advanced human-like autonomous races, a contextual understanding based on learning methods plays a key role. We believe this technology has an enormous impact in multiple application domains ranging from urban autonomous driving to service robots in social environments. At the same time, we would like to leave an open research question: \emph{"How can we maintain the system resilience when adopting unobservable learning models?"}. One promising research topic is explainable deep learning, which has evolved significantly in the last few years. Also, system designs that combine classical and learning-based methods in a complementary way can represent a solid solution to the problem of deployable learning models. With efforts in these fields, we believe that learning algorithms can be more actively adopted in real-world autonomous systems.

\section*{Acknowledgments}
This work is partially supported by SK Hynix Inc and Institute of Information communications Technology Planning Evaluation (IITP) grant funded by the Korea government (MSIT, 2021-0-00029). We would like to thank Energy System Network (ESN), Juncos Hollinger Racing, and all other participating teams for their support and contributions to the project. Especially, we would like to thank \emph{MIT-PITT-RW} for collaborating on an LQR controller design and camera-based perception module. Also, we thank \emph{PoliMOVE} for sharing their semi-final log data with us, which was used for perception performance evaluation in this work.

\bibliographystyle{apacite}
\bibliography{frExample}

\end{document}